\documentclass[11pt,a4paper]{article}
\usepackage[hyperref]{emnlp2020}
\usepackage{times}
\usepackage{latexsym}

\usepackage{graphicx}
\usepackage[export]{adjustbox}
\usepackage{url}
\usepackage{bm}
\usepackage{here}
\usepackage{amsmath}
\usepackage{amssymb}
\usepackage{amsfonts}
\usepackage{multirow}
\usepackage{arydshln}
\usepackage{enumitem}

\usepackage{algorithm}
\usepackage{algpseudocode}
\usepackage{soul}
\usepackage{float}
\usepackage{subfigure}
\usepackage{microtype}

\aclfinalcopy 

\newcommand{\argmax}[1]{\underset{#1}{\operatorname{arg}\,\operatorname{max}}\;}

\title{Discriminative Nearest Neighbor Few-Shot Intent Detection\\ by Transferring Natural Language Inference}

\author{
  \textbf{Jian-Guo Zhang}$^1$\thanks{\ \ Work done while the first author was an intern at Salesforce Research.}~~~~ \textbf{Kazuma Hashimoto}$^2$\thanks{\ \ Corresponding author.}~~~~ \textbf{Wenhao Liu}$^2$~~~~ \textbf{Chien-Sheng Wu}$^2$~~~~  \\ \textbf{Yao Wan}$^3$~~~~ \textbf{Philip S. Yu}$^1$~~~~  \textbf{Richard Socher}$^2$~~~~ \textbf{Caiming Xiong}$^2$ \\
  $^1$ University of Illinois at Chicago, Chicago, USA \\
  $^2$Salesforce Research,  Palo Alto, USA \\
  $^3$Huazhong University of Science and Technology, Wuhan, China\\
  { \texttt{\{jzhan51,psyu\}@uic.edu}, \texttt{wanyao@hust.edu.cn} } \\
  { \texttt{\{{k.hashimoto,wenhao.liu,wu.jason,rsocher,cxiong\}@salesforce.com}} }
}

\date{}

\begin{document}
\maketitle
\begin{abstract}



Intent detection is one of the core components of goal-oriented dialog systems, and detecting out-of-scope (OOS) intents is also a practically important skill.
Few-shot learning is attracting much attention to mitigate data scarcity, but OOS detection becomes even more challenging.
In this paper, we present a simple yet effective approach, discriminative nearest neighbor classification with deep self-attention.
Unlike softmax classifiers, we leverage BERT-style pairwise encoding to train a binary classifier that estimates the best matched training example for a user input.
We propose to boost the discriminative ability by transferring a natural language inference (NLI) model.
Our extensive experiments on a large-scale multi-domain intent detection task show that our method achieves more stable and accurate in-domain and OOS detection accuracy than RoBERTa-based classifiers and embedding-based nearest neighbor approaches.
More notably, the NLI transfer enables our 10-shot model to perform competitively with 50-shot or even full-shot classifiers, while we can keep the inference time constant by leveraging a faster embedding retrieval model. 

\end{abstract}

\section{Introduction}

\begin{figure}[ht]
	\begin{center}
    	\includegraphics[width=1.0\linewidth]{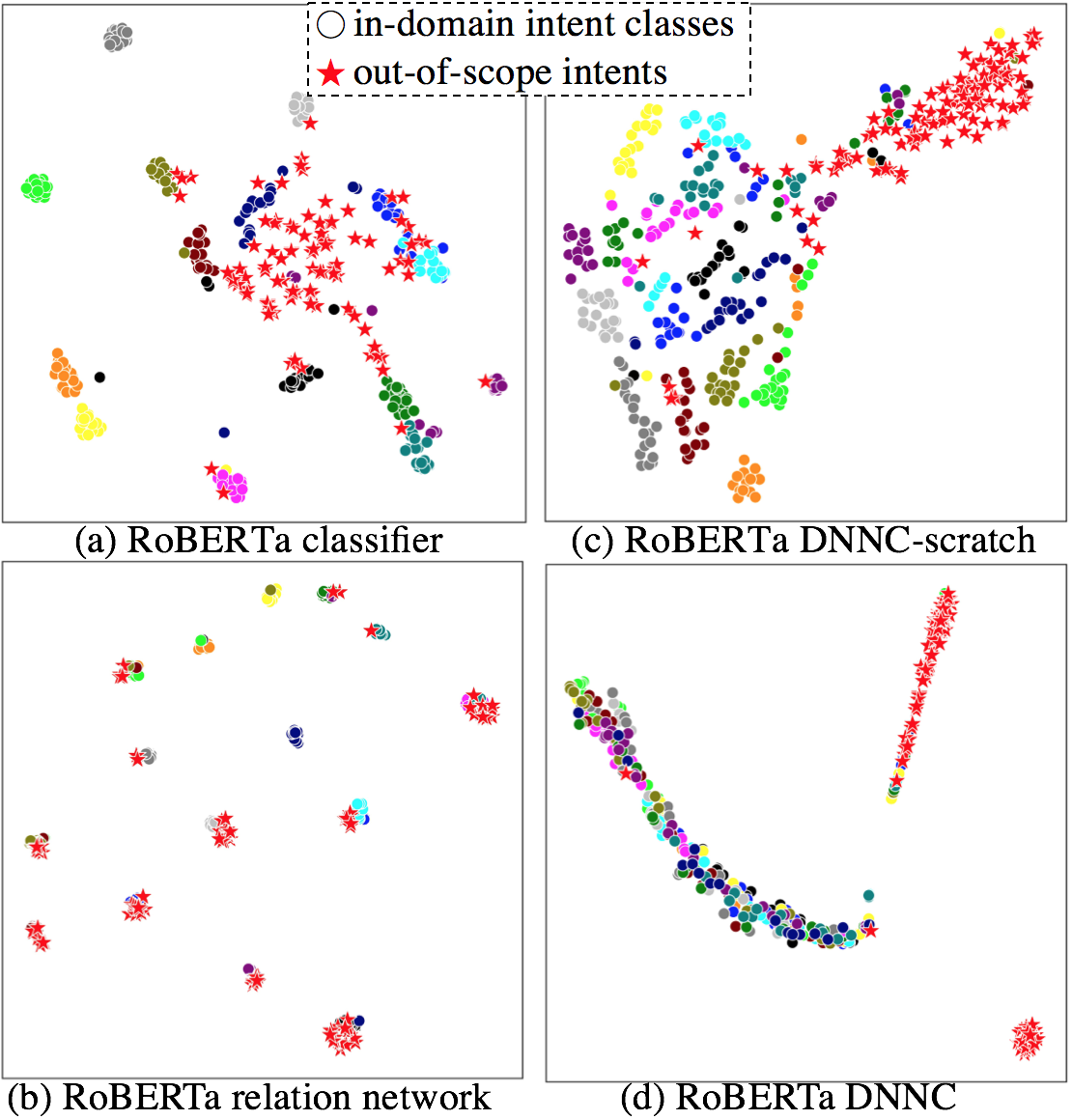}
    \end{center}
\caption{tSNE visual comparison for OOS detection between existing methods ((a) and (b)) and our proposed method ((c) and (d)). Their embeddings before their classifier layers are used (best viewed in color).}
\label{fig:tsne-main}
\vspace{-1em}
\end{figure}

Intent detection is one of the core components when building goal-oriented dialog systems.
The goal is to achieve high intent classification accuracy, and another important skill is to accurately detect unconstrained user intents that are out-of-scope (OOS) in a system~\citep{oos-intent}.
A practical challenge is data scarcity because different systems define different sets of intents, and thus few-shot learning is attracting much attention.
However, previous work has mainly focused on the few-shot intent classification without OOS~\citep{marrying-reg,efficient_intent}.

OOS detection can be considered as out-of-distribution detection~\citep{softmax_conf,learning_ood}.
Recent work has shown that large-scale pre-trained models like BERT~\citep{bert} and RoBERTa~\citep{roberta} still struggle with out-of-distribution detection, despite their strong in-domain performance~\citep{acl-ood}.
Figure~\ref{fig:tsne-main} (a) shows how unseen input text is mapped into a feature space, by a RoBERTa-based model for 15-way 5-shot intent classification.
The separation between OOS and some in-domain intents is not clear, which presumably hinders the model's OOS detection ability.
This observation calls for investigation into more sample-efficient approaches to handling the in-domain and OOS examples accurately.

In this paper, we tackle the task from a different angle, and propose a discriminative nearest neighbor classification (DNNC) model.
Instead of expecting the text encoders to be generalized enough to discriminate both the in-domain and OOS examples, we make full use of the limited training examples both in training and inference time as a nearest neighbor classification schema.
We leverage the BERT-style paired text encoding with deep self-attention to directly model relations between pairs of user utterances.
We then train a matching model as a pairwise binary classifier to estimate whether an input utterance belongs to the same class of a paired example.
We expect this to free the model from having the OOS separation issue in Figure~\ref{fig:tsne-main} (a) by avoiding explicit modeling of the intent classes.
Unlike an embedding-based matching function as in relation networks~\citep{relation_net} (Figure~\ref{fig:tsne-main} (b)), the deep pairwise matching function produces clear separation between the in-domain and OOS examples (Figure~\ref{fig:tsne-main} (c)).
We further propose to seamlessly transfer a natural language inference (NLI) model to enhance this clear separation (Figure~\ref{fig:tsne-main} (d)).

We verify our hypothesis by conducting extensive experiments on a large-scale multi-domain intent detection task with OOS~\citep{oos-intent} in various few-shot learning settings.
Our experimental results show that, compared with RoBERTa classifiers and embedding nearest neighbor approaches, our DNNC attains more stable and accurate performance both in in-domain and OOS accuracy.
Moreover, our 10-shot model can perform competitively with a 50-shot or even full-shot classifier, with the performance boost by the NLI transfer.
We also show how to speedup our DNNC's inference time without sacrificing accuracy.

\section{Background}


\subsection{Task: Few-Shot Intent Detection}
\label{subsec:task-def}

Given a user utterance $u$ at every turn in a goal-oriented dialog system, an intent detection model $I(u)$ aims at predicting the speaker's intent:
\begin{equation}
    I(u) = c,
\end{equation}
where $c$ is one of pre-defined $N$ intent classes $\mathbf{C}=\{C_1, C_2, \ldots, C_N\}$, or is categorized as OOS.
The OOS category corresponds to user utterances whose requests are not covered by the system.
In other words, any utterance can be OOS as long as it does {\it not} fall into any of the $N$ intent classes, so the definition of OOS is different depending on $\mathbf{C}$.

\paragraph{Balanced $K$-shot learning}
In a few-shot learning scenario, we have a limited number of training examples for each class, and we assume that we have $K$ examples for each of the $N$ classes in our training data.
In other words, we have $N\cdot K$ training examples in total.
We denote the $i$-th training example from the $j$-th class $C_j$ as $e_{j,i}\in E$, where $E$ is the set of the examples.
$K$ is typically 5 or 10.

\subsection{Multi-Class Classification}
\label{subsec:classifier}

The goal is to achieve high accuracy both for the intent classification and OOS detection.
One common approach to this task is using a multi-class classification model.
Specifically, to get a strong baseline for the few-shot learning use case, one can leverage a pre-trained model as transfer learning, which has been shown to achieve state-of-the-art results on numerous natural language processing tasks.
We use BERT~\citep{bert,roberta} as a text encoder:
\begin{equation}
    \label{eq:bert}
    h = \mathrm{BERT}(u) \in\mathbb{R}^{d},
\end{equation}
where $h$ is a $d$-dimensional output vector corresponding to the special token {\tt[CLS]} as in the follwing input format: [{\tt[CLS]}, $u$, {\tt[SEP]}].\footnote{The format of these special tokens is different in RoBERTa, but we use the original BERT's notations.}

To handle the intent classification and the OOS detection, we apply the threshold-based strategy in \citet{oos-intent}, to the softmax output of the $N$-class classification model~\citep{softmax_conf}:
\begin{equation}
    \label{eq:softmax}
    p(c | u) = \mathrm{softmax}(W h + b) \in\mathbb{R}^{N},
\end{equation}
where $W\in\mathbb{R}^{N\times d}$ and $b\in\mathbb{R}^{N}$ are the classifier's model parameters.
The classification model is trained by cross-entropy loss with the ground-truth intent labels of the training examples.
At inference time, we first take the class $C_j$ with the largest value of $p(c=C_j|u)$, then output $I(u)=C_j$ if $p(c=C_j|u) > T$, where $T\in[0.0, 1.0]$ is a threshold to be tuned, and otherwise we output $I(u)=\mathrm{OOS}$.

\begin{table*}[t]

 \begin{center}
{\small
 \resizebox{1.0\linewidth}{!}{
    \begin{tabular}{l|l|l|l}
    
    & Input utterance & Utterance to be compared & Label \\ \hline
    (a) & i need you to send 500 dollars from & help me move my money please & pos. \\
        & my high tier account to my regular checking & & \\ \hdashline
    (b) & i would like to know when the bill is due & i need to know the amounts due for my utilities and cable bills & neg. \\ \hline

    (c) & And can you tell me any of the names and & Are you able to inform me of any name or address? & pos. \\
        & addresses? Annie considered. & & \\ \hdashline
    (d) & It's Sunday, what channel is this? & It's Sunday, can you change the channel? & neg. \\ \hdashline
    (e) & I want to go back to Marguerite. & I never want to return to Marguerite. & neg. \\ \hline

    \end{tabular}}
    \caption{Training examples for our model. The first two examples ((a)--(b)) come from the CLINC150 dataset~\citep{oos-intent}, and the other three examples ((c)--(e)) come from the MNLI dataset~\citep{mnli}.}
    \label{tb:examples}
    \vspace{-1em}
}
  \end{center}
\end{table*}

\subsection{Nearest Neighbor Classification}

As the fundamental building block of our proposed method, we also review nearest neighbor classification (i.e., $k$-nearest neighbors ($k$NN) classification with $k=1$), a simple and well-established concept for classification~\citep{knn-nips,knn-classifier}.
The basic idea is to classify an input into the same class of the most relevant training example based on a certain metric.

In our task, we formulate a nearest neighbor classification model as the following:
\begin{equation}
    \label{eq:example-based}
    I(u) = \mathrm{class}\left(\argmax{e_{j, i}\in E} S(u, e_{j, i}) \right),
\end{equation}
where $\mathrm{class}(e_{j, i})$ is a function returning the intent label (class) of the training example $e_{j, i}$, and $S$ is a function that estimates some relevance score between $u$ and $e_{j, i}$.
To detect OOS, we can also use the uncertainty-based strategy in Section~\ref{subsec:classifier}; that is, we take the output label from Equation~(\ref{eq:example-based}) if the corresponding relevance score is greater than a threshold, and otherwise we output $I(u)=\mathrm{OOS}$.

\section{Proposed Method}
\label{subsec:method}

This section first describes how to directly model inter-utterance relations in our nearest neighbor classification scenario.
We then introduce a binary classification strategy by synthesizing pairwise examples, and propose a seamless transfer of NLI.
Finally, we describe how to speedup our method's inference process.

\subsection{Deep Pairwise Matching Function}
The objective of $S(u, e_{j, i})$ in Equation~(\ref{eq:example-based}) is to find the best matched utterance from the training set $E$, given the input utterance $u$.
The typical methodology is to embed each data example into a vector space and (1) use an off-the-shelf distance metric to perform a similarity search~\citep{knn-classifier} or (2) learn a distant metric between the embeddings~\citep{relation_net}.
However, as shown in Figure~\ref{fig:tsne-main}, the text embedding methods do not discriminate the OOS examples well enough.

To model fine-grained relations of utterance pairs to distinguish in-domain and OOS intents, we propose to formulate $S(u, e_{j, i})$ as follows:
\begin{eqnarray}
    h &=& \mathrm{BERT}(u, e_{j, i}) \in\mathbb{R}^{d}, \label{eq:NLI} \\
    S(u, e_{j, i}) &=& \sigma(W \cdot h + b) \in\mathbb{R}, \label{eq:sigmoid}
\end{eqnarray}
where $\mathrm{BERT}$ is the same model in Equation~(\ref{eq:bert}), except that we follow a different input format to accommodate pairs of utterances: [{\tt[CLS]}, $u$, {\tt[SEP]}, $e_{j, i}$, {\tt[SEP]}].
$\sigma$ is the sigmoid function, and $W\in\mathbb{R}^{1\times d}$ and $b\in\mathbb{R}$ are the model parameters.
We can interpret our method as wrapping both the embedding and matching functions into the paired encoding with the deep self-attention in BERT (Equation~(\ref{eq:NLI})) along with the discriminative model (Equation~(\ref{eq:sigmoid})).
It has been shown that the paired text encoding is crucial in capturing complex relations between queries and documents in document retrieval~\citep{watanabe-query,sem-bert,graph-retriever}.

\subsection{Discriminative Training}
\label{subsec:pos_neg}

We train the matching model $S(u, e_{j, i})$ as a binary classifier, such that $S(u, e_{j, i})$ is closed to 1.0 if $u$ belongs to the same class of $e_{j, i}$, and otherwise closed to 0.0.
The model is trained by a binary cross-entropy loss function.

\paragraph{Positive examples}
To create positive examples, we consider all the possible ordered pairs within the same intent class: $(e_{j, i}, e_{j, \ell})$ such that $i \neq \ell$.
We thus have $N\times K\times(K-1)$ positive examples in total.
Table~\ref{tb:examples} (a) shows a positive example created from an intent, ``transfer,'' in a banking domain.

\paragraph{Negative examples}
For negative examples, we consider all the possible ordered pairs across any two different intent classes: $(e_{j, i}, e_{o, \ell})$ such that $j \neq o$.
We thus have $K^2\times N\times(N-1)$ negative examples in total, and this number is in general greater than that of the positive examples.
Table~\ref{tb:examples} (b) shows a negative example, where the input utterance comes from the intent, ``bill due'', and the paired sentence from another intent, ``bill balance''.

\subsection{Seamless Transfer from NLI}
\label{subsec:nli_transfer}

A key characteristic of our method is that we seek to model the relations between the utterance pairs, instead of explicitly modeling the intent classes.
To mitigate the data scarcity setting in few-shot learning, we consider transferring another inter-sentence-relation task.

This work focuses on NLI; the task is to identify whether a hypothesis sentence can be entailed by a premise sentence~\citep{deep-nli}.
We treat the NLI task as a binary classification task: {\tt entailment} (positive) or {\tt non-entailment} (negative).\footnote{A widely-used format is a three-way classification task with {\tt entailment}, {\tt neutral}, and {\tt contradiction}, but we merge the latter two classes into a single {\tt non-entailment} class.}
We first pre-train our model with the NLI task, where the premise sentence corresponds to the $u$-position, and the hypothesis sentence corresponds to the $e_{j, i}$-position in Equation~(\ref{eq:NLI}).
Note that it is not necessary to modify the model architecture since the task format is consistent, and we can train the NLI model solely based on existing NLI datasets.
Once the NLI model pre-training is completed, we fine-tune the NLI model with the intent classification training examples described in Section~\ref{subsec:pos_neg}.
This allows us to transfer the NLI model to any intent detection datasets seamlessly.

\paragraph{Why NLI?}
The NLI task has been actively studied, especially since the emergence of large scale datasets~\citep{snli,mnli}, and we can directly leverage the progress.
Moreover, recent work is investigating cross-lingual NLI~\citep{eriguchi-crosslingual,xnli}, and this is encouraging to consider multilinguality in future work.
On the other hand, while we can find examples relevant to the intent detection task, as shown in Table~\ref{tb:examples} ((c), (d), and (e)), we still need the few-shot fine-tuning.
This is because a domain mismatch still exists in general, and perhaps more importantly, our intent detection approach is not exactly modeling NLI.

\paragraph{Why not other tasks?}
There are other tasks modeling relationships between sentences.
Paraphrase~\citep{paranmt} and semantic relatedness~\citep{semeval} tasks are such examples.
It is possible to automatically create large-scale paraphrase datasets by machine translation~\citep{ppdb}.
However, our task is not a paraphrasing task, and creating negative examples is crucial and non-trivial~\citep{selectional-preference}.
In contrast, as described above, the NLI setting comes with negative examples by nature.
The semantic relatedness (or textual similarity) task is considered as a coarse-grained task compared to NLI, as discussed in the previous work~\citep{jmt}, in that the task measures semantic or topical relatedness.
This is not ideal for the intent detection task, because we need to discriminate between topically similar utterances of different intents.
In summary, the NLI task well matches our objective, with access to the large datasets.

\subsection{A Joint Approach with Fast Retrieval}
\label{subsec:joint}

The number of model parameters of the multi-class classification model in Section~\ref{subsec:classifier} and our model in Section~\ref{subsec:method} is almost the same when we use the same pre-trained models.
However, our example-based method has an inference-time bottleneck in Equation~(\ref{eq:NLI}), where we need to compute the BERT encoding for all $N\times K$ $(u, e_{j,i})$ pairs. 

We follow common practice in document retrieval to reduce the inference-time bottleneck~\citep{sem-bert,graph-retriever}, by introducing a fast text retrieval model to select a set of top-$k$ examples $E_k$ from the training set $E$, based on its retrieval scores.
We then replace $E$ in Equation~(\ref{eq:example-based}) with the shrunk set $E_k$.
The cost of the paired BERT encoding is now constant, regardless the size of $E$.
Either TF-IDF~\citep{drqa} or embedding-based retrieval~\citep{faiss,phrase-embedding-index,latent-retrieval} can be used for the first step.
We use the following fast $k$NN.

\paragraph{Faster $k$NN}
As a baseline and a way to instantiate our joint approach, we use Sentence-BERT (SBERT)~\citep{sbert} to separately encode $u$ and $e_{j, i}$ ($x\in\{u, e_{j, i}\}$) as follows:
\begin{eqnarray}
    \label{eq:sbert}
    v(x) &=& \mathrm{SBERT}(x) \in\mathbb{R}^{d},
\end{eqnarray}
where the input text format is identical to that of BERT in Equation~(\ref{eq:bert}).
SBERT is a BERT-based text embedding model, fine-tuned by siamese networks with NLI datasets.
Thus both our method and SBERT transfer the NLI task in different ways.

Cosine similarity between $v(u)$ and $v(e_{j, i})$ then replaces $S(u, e_{j, i})$ in Equation~(\ref{eq:sigmoid}).
To get a fair comparison, instead of using the encoding vectors produced by the original SBERT, we fine-tune SBERT with our intent training examples described in Section~\ref{subsec:pos_neg}.
The cosine similarity is symmetric, so we have half the training examples.
We use the pairwise cosine-based loss function in \citet{sbert}.
After the model training, we pre-compute $v(e_{j, i})$ for fast retrieval.



\section{Experimental Settings}

\subsection{Dataset: Multi-Domain Intent Detection}

\begin{table}[t]
  \begin{center}
{\small
    \begin{tabular}{l|r||r|r|r}

                  & $N$ & Train & Dev. & Test \\ \hline
    All domains   & 150 & 15,000 & 3,000 & 4,500 \\
    Single domain & 15  & 1,500  & 300   & 450 \\
    OOS           & -   & n/a    & 100   & 1,000 \\ \hline

    \end{tabular}
}
    \caption{Dataset statistics. The number of the examples is equally distributed across the intent classes.}
    \label{tb:dataset}
    \vspace{-1em}
  \end{center}
\end{table}

We use a recently-released dataset, CLINC150,\footnote{\url{https://github.com/clinc/oos-eval}.} for multi-domain intent detection~\citep{oos-intent}.
The CLINC150 dataset defines 150 types of intents in total (i.e., $N=150$), where there are 10 different domains and 15 intents for each of them.
Table~\ref{tb:dataset} shows the dataset statistics.

\paragraph{OOS evaluation examples}
The dataset also provides OOS examples whose intents do not belong to any of the 150 intents.
From the viewpoint of out-of-distribution detection~\citep{softmax_conf,acl-ood}, we do not use the OOS examples during the training stage; we only use the evaluation splits as in Table~\ref{tb:dataset}.

\paragraph{Single-domain experiments}
The task in the CLINC150 dataset is like handling many different services in a single system; that is, topically different intents are mixed (e.g., ``alarm'' in the ``Utility'' domain, and ``pay bill'' in the ``Banking'' domain).
In contrast, it is also a reasonable setting to handle each domain (or service) separately as in \citet{google-schema-guided}.
In addition to the all-domain experiment, we conduct single-domain experiments, where we only focus on a specific domain with its 15 intents (i.e., $N=15$).
More specifically, we use four domains, ``Banking,'' ``Credit cards,'' ``Work,'' and ``Travel,'' among the ten domains.
Note that the same OOS evaluation sets are used.

\subsection{Evaluation Metrics}
\label{subsection:evalMetrics}

We follow \citet{oos-intent} to report in-domain accuracy, $\mathrm{Acc}_\mathrm{in}$, and OOS recall, $R_\mathrm{oos}$.
These two metrics are defined as follows:
\begin{equation}
        \label{eq:two_metrics}
        \mathrm{Acc}_\mathrm{in} = C_\mathrm{in}/N_\mathrm{in}, ~~~ R_\mathrm{oos} = C_\mathrm{oos}/N_\mathrm{oos}, \\
\end{equation}
where $C_\mathrm{in}$ is the number of correctly predicted in-domain intent examples, and $N_\mathrm{in}$ is the total number of the examples evaluated.
This is analogous to the calculation of the OOS recall.

\paragraph{Threshold selection}
We use the uncertainty-based OOS detection, and therefore we need a way to set the threshold $T$.
For each $T$ in $[0.0, 0.1, \ldots, 0.9, 1.0]$, we calculate a joint score $J_\mathrm{in\_oos}$ defined as follows:
\begin{equation}
\label{eq:joint_score}
J_\mathrm{in\_oos} = \mathrm{Acc}_\mathrm{in} + R_\mathrm{oos},
\end{equation}
and select a threshold value to maximize the score on the development set.
There is a trade-off to be noted; the larger the value of $T$ is, the higher $R_\mathrm{oos}$ (and the lower $\mathrm{Acc}_\mathrm{in}$) we expect, because the models predict OOS more frequently.

\paragraph{Notes on $J_\mathrm{in\_oos}$}
Our joint score $J_\mathrm{in\_oos}$ in Equation~(\ref{eq:joint_score}) gives the same weight to the two metrics, $\mathrm{Acc}_\mathrm{in}$ and $R_\mathrm{oos}$, compared to other combined metrics.
For example, \citet{oos-intent} and \citet{tod-bert} used a joint accuracy score: 
\begin{equation}
\frac{C_\mathrm{in}+C_\mathrm{oos}}{N_\mathrm{in}+N_\mathrm{oos}} = \frac{\mathrm{Acc}_\mathrm{in} + r R_\mathrm{oos}}{1 + r},
\end{equation}
where $r=N_\mathrm{oos}/N_\mathrm{in}$ depends on the balance between $N_\mathrm{in}$ and $N_\mathrm{oos}$, and thus this combined metric can put much more weight on the in-domain accuracy when $N_\mathrm{in} \gg N_\mathrm{oos}$.
Table~\ref{tb:dataset} shows $r=100/3000~(=0.0333)$ in the development set of the ``all domains'' setting, which underestimates the importance of $R_\mathrm{oos}$.
Actually, the OOS recall scores in \citet{oos-intent} and \citet{tod-bert} are much lower than those with our RoBERTa classifier, and the trade-off with respect to the tuning process was not discussed.\footnote{When we refer to our RoBERTa classifier's scores $(A_\mathrm{in}, R_\mathrm{oos})=(92.9\pm0.2,~90.2\pm0.5)$ in Table~\ref{table:wholeDataset}, their corresponding scores are $(96.2,~52.3)$ and $(96.1,~46.3)$ in \citet{oos-intent} and \citet{tod-bert}, respectively.}

We also report OOS precision and OOS F1 for more comprehensive evaluation:
\begin{equation}
    P_\mathrm{oos} = C_\mathrm{oos}/N'_\mathrm{oos}, ~~~ F_1 = H(P_\mathrm{oos}, R_\mathrm{oos}),
\end{equation}
where $N'_\mathrm{oos}$ is the number of examples predicted as OOS, and $H(\cdot, \cdot)$ is the harmonic mean.

\subsection{Model Training and Configurations}
\label{subsec:model_setting}

We use RoBERTa (the {\tt base} configuration with $d=768$) as a BERT encoder for all the BERT/SBERT-based models in our experiments,\footnote{We use \url{https://github.com/huggingface/transformers} and \url{https://github.com/UKPLab/sentence-transformers}.}
because RoBERTa performed significantly better and more stably than the original BERT in our few-shot experiments.
We combine three NLI datasets, SNLI~\citep{snli}, MNLI~\citep{mnli}, and WNLI~\citep{wnli} from the GLUE benchmark~\citep{glue} to pre-train our proposed model.

We apply label smoothing~\citep{label_smoothing_orig} to all the cross-entropy loss functions, which has been shown to improve the reliability of the model confidence~\citep{label_smoothing}.
Experiments were conducted on single NVIDIA Tesla V100 GPU with 16GB memory.

\paragraph{Sampling training examples}
We conduct our experiments with $K=5, 10$ following the task definition in Section~\ref{subsec:task-def}.
We randomly sample $K$ examples from the entire training sets in Table~\ref{tb:dataset}, for each in-domain intent class 10 times unless otherwise stated.
We train a model with a consistent hyper-parameter setting across the 10 different runs and follow the threshold selection process based on a mean score for each threshold.
We also report a standard deviation for each result.

\begin{table*}[h]
\centering
\tiny
\resizebox{1.0\linewidth}{!}{
\begin{tabular}{lcccccccc}
\hline
\multicolumn{1}{l|}{}                 & \multicolumn{2}{c|}{\textbf{In-domain accuracy}}                                                 & \multicolumn{2}{c|}{\textbf{OOS recall}}                                                         & \multicolumn{2}{c|}{\textbf{OOS precision}}                                                      & \multicolumn{2}{c}{\textbf{OOS F1}}                                         \\
\multicolumn{1}{l|}{\textbf{5-shot}}  & \textbf{Banking}                     & \multicolumn{1}{c|}{\textbf{Credit cards}}               & \textbf{Banking}                     & \multicolumn{1}{c|}{\textbf{Credit cards}}               & \textbf{Banking}                     & \multicolumn{1}{c|}{\textbf{Credit cards}}               & \textbf{Banking}                     & \textbf{Credit cards}               \\ \hline
\multicolumn{1}{l|}{classifier}       & 79.7 $\pm$ 1.8          & \multicolumn{1}{c|}{79.9 $\pm$ 3.7}          & 93.3 $\pm$ 2.9          & \multicolumn{1}{c|}{93.3 $\pm$ 3.0}          & 92.5  $\pm$ 0.9         & \multicolumn{1}{c|}{93.6 $\pm$ 1.3}          & 92.9 $\pm$ 1.8          & 93.4 $\pm$ 1.6          \\
\multicolumn{1}{l|}{classifier-EDA}   & 79.9 $\pm$ 3.0          & \multicolumn{1}{c|}{73.1 $\pm$ 5.0}          & 91.0 $\pm$ 2.6          & \multicolumn{1}{c|}{94.1 $\pm$ 2.6}          & 92.6  $\pm$ 1.5         & \multicolumn{1}{c|}{90.0 $\pm$ 1.7}          & 91.8 $\pm$ 1.9          & 92.0 $\pm$ 1.3          \\
\multicolumn{1}{l|}{classifier-BT}    & 77.2  $\pm$ 2.8         & \multicolumn{1}{c|}{70.3 $\pm$ 4.4}          & 91.5 $\pm$ 2.4          & \multicolumn{1}{c|}{91.1 $\pm$ 1.8}          & 91.4 $\pm$ 1.2          & \multicolumn{1}{c|}{89.5 $\pm$ 1.8}          & 91.4 $\pm$ 1.4          & 90.3 $\pm$ 1.7          \\
\multicolumn{1}{l|}{DNNC-scratch}      & 83.4 $\pm$ 1.9          & \multicolumn{1}{c|}{84.6 $\pm$ 3.2}          & 93.2 $\pm$ 2.7          & \multicolumn{1}{c|}{96.4 $\pm$ 1.0}          & 96.2 $\pm$ 0.9          & \multicolumn{1}{c|}{95.8 $\pm$ 1.1}          & 94.7 $\pm$ 1.2          & 96.1 $\pm$ 0.9          \\
\multicolumn{1}{l|}{DNNC}    & \textbf{88.6 $\pm$ 1.3} & \multicolumn{1}{c|}{\textbf{90.5 $\pm$ 2.9}} & \textbf{94.7 $\pm$ 1.0} & \multicolumn{1}{c|}{\textbf{97.7 $\pm$ 1.1}} & \textbf{97.0 $\pm$ 0.3} & \multicolumn{1}{c|}{\textbf{97.8 $\pm$ 0.5}} & \textbf{95.9 $\pm$ 0.5} & \textbf{97.8 $\pm$ 0.6} \\ \hdashline
\multicolumn{1}{l|}{classifier (50-shot)}       &   90.0 $\pm$ 1.4       & \multicolumn{1}{c|}{90.3 $\pm$ 1.4}          &     93.3 $\pm$ 1.1     & \multicolumn{1}{c|}{93.9 $\pm$ 1.2}          &     96.3 $\pm$ 0.6     & \multicolumn{1}{c|}{96.4 $\pm$ 0.6}          &    94.8  $\pm$ 0.7     &  95.1 $\pm$ 0.6        \\ \hline
\multicolumn{1}{l|}{\textbf{5-shot}}  & \textbf{Work}                        & \multicolumn{1}{c|}{\textbf{Travel}}                      & \textbf{Work}                        & \multicolumn{1}{c|}{\textbf{Travel}}                      & \textbf{Work}                        & \multicolumn{1}{c|}{\textbf{Travel}}                      & \textbf{Work}                        & \textbf{Travel}                      \\ \hline
\multicolumn{1}{l|}{classifier}       & 84.4 $\pm$ 1.9          & \multicolumn{1}{c|}{87.8 $\pm$ 2.5}          & 94.8 $\pm$ 1.6          & \multicolumn{1}{c|}{96.1 $\pm$ 0.9}          & 95.4 $\pm$ 0.7          & \multicolumn{1}{c|}{94.7 $\pm$ 1.0}          & 95.1 $\pm$ 0.9          & 95.4 $\pm$ 0.7          \\
\multicolumn{1}{l|}{classifier-EDA}   & 80.5 $\pm$ 3.3          & \multicolumn{1}{c|}{88.2 $\pm$ 2.4}          & 95.2 $\pm$ 1.6          & \multicolumn{1}{c|}{94.6 $\pm$ 2.1}          & 92.8 $\pm$ 1.3          & \multicolumn{1}{c|}{94.9 $\pm$ 1.0}          & 94.0 $\pm$ 0.9          & 94.7 $\pm$ 0.9          \\
\multicolumn{1}{l|}{classifier-BT}    & 80.3 $\pm$ 3.4          & \multicolumn{1}{c|}{90.8 $\pm$ 2.4}          & 94.4 $\pm$ 1.1          & \multicolumn{1}{c|}{92.3 $\pm$ 1.9}          & 93.1 $\pm$ 1.0          & \multicolumn{1}{c|}{95.9 $\pm$ 1.1}          & 93.8 $\pm$ 0.8          & 94.1 $\pm$ 1.3          \\
\multicolumn{1}{l|}{DNNC-scratch}      & 83.2 $\pm$ 2.1          & \multicolumn{1}{c|}{87.8 $\pm$ 3.5}          & 96.3 $\pm$ 1.7          & \multicolumn{1}{c|}{94.9 $\pm$ 2.8}          & 96.7 $\pm$ 0.9          & \multicolumn{1}{c|}{94.9 $\pm$ 1.4}          & 96.5 $\pm$ 0.6          & 94.9 $\pm$ 0.9          \\
\multicolumn{1}{l|}{DNNC}    & \textbf{89.9 $\pm$ 3.2} & \multicolumn{1}{c|}{\textbf{91.8 $\pm$ 1.6}} & \textbf{96.7 $\pm$ 1.1} & \multicolumn{1}{c|}{\textbf{96.4 $\pm$ 1.2}} & \textbf{97.9 $\pm$ 1.1} & \multicolumn{1}{c|}{\textbf{96.5 $\pm$ 0.7}} & \textbf{97.3 $\pm$ 0.5} & \textbf{96.5 $\pm$ 0.7} \\ \hdashline
\multicolumn{1}{l|}{classifier (50-shot)}       &  94.3 $\pm$ 0.8      & \multicolumn{1}{c|}{97.0 $\pm$ 0.3}          &   95.2 $\pm$ 0.8       & \multicolumn{1}{c|}{92.6 $\pm$ 1.4}          &     97.7 $\pm$ 0.3     & \multicolumn{1}{c|}{98.6 $\pm$ 0.2}          &     96.5 $\pm$ 0.5     &    95.5 $\pm$ 0.8      \\ \hline
                                      & \multicolumn{1}{l}{}                 & \multicolumn{1}{l}{}                                      & \multicolumn{1}{l}{}                 & \multicolumn{1}{l}{}                                      & \multicolumn{1}{l}{}                 & \multicolumn{1}{l}{}                                      & \textbf{}                            & \textbf{}                            \\ \hline
\multicolumn{1}{l|}{}                 & \multicolumn{2}{c|}{\textbf{In-domain accuracy}}                                                 & \multicolumn{2}{c|}{\textbf{OOS recall}}                                                         & \multicolumn{2}{c|}{\textbf{OOS precision}}                                                      & \multicolumn{2}{c}{\textbf{OOS F1}}                                         \\
\multicolumn{1}{l|}{\textbf{10-shot}} & \textbf{Banking}                     & \multicolumn{1}{c|}{\textbf{Credit cards}}               & \textbf{Banking}                     & \multicolumn{1}{c|}{\textbf{Credit cards}}               & \textbf{Banking}                     & \multicolumn{1}{c|}{\textbf{Credit cards}}               & \textbf{Banking}                     & \textbf{Credit cards}               \\ \hline
\multicolumn{1}{l|}{classifier}       & 85.2 $\pm$ 1.3          & \multicolumn{1}{c|}{83.7 $\pm$ 2.1}          & 93.3 $\pm$ 1.0          & \multicolumn{1}{c|}{93.8 $\pm$ 1.4}          & 94.4 $\pm$ 0.5          & \multicolumn{1}{c|}{93.9 $\pm$ 0.9}          & 93.8 $\pm$ 0.6          & 93.8 $\pm$ 0.7          \\
\multicolumn{1}{l|}{classifier-EDA}   & 82.5 $\pm$ 1.3          & \multicolumn{1}{c|}{79.3 $\pm$ 1.7}          & \textbf{95.9 $\pm$ 0.8} & \multicolumn{1}{c|}{96.8 $\pm$ 0.8}          & 93.3 $\pm$ 0.4          & \multicolumn{1}{c|}{92.3 $\pm$ 0.8}          & 94.6 $\pm$ 0.5          & 94.5 $\pm$ 0.3          \\
\multicolumn{1}{l|}{classifier-BT}    & 82.2 $\pm$ 1.9          & \multicolumn{1}{c|}{82.9 $\pm$ 1.8}          & 94.9 $\pm$ 1.9          & \multicolumn{1}{c|}{89.3 $\pm$ 2.0}          & 93.1 $\pm$ 0.8          & \multicolumn{1}{c|}{94.3 $\pm$ 0.6}          & 94.0 $\pm$ 0.9         & 91.7 $\pm$ 1.2          \\
\multicolumn{1}{l|}{DNNC-scratch}      & 89.6 $\pm$ 1.6          & \multicolumn{1}{c|}{92.1 $\pm$ 1.1}          & 92.1 $\pm$ 3.1          & \multicolumn{1}{c|}{94.8 $\pm$ 1.2}          & 97.5 $\pm$ 0.9          & \multicolumn{1}{c|}{\textbf{98.1 $\pm$ 0.4}} & 94.7 $\pm$ 1.5          & 96.4 $\pm$ 0.6          \\
\multicolumn{1}{l|}{DNNC}    & \textbf{91.2 $\pm$ 1.1} & \multicolumn{1}{c|}{\textbf{92.1 $\pm$ 1.0}} & 94.8 $\pm$ 1.1          & \multicolumn{1}{c|}{\textbf{97.8 $\pm$ 0.8}} & \textbf{97.5 $\pm$ 0.4} & \multicolumn{1}{c|}{97.8 $\pm$ 0.3}          & \textbf{96.1 $\pm$ 0.6} & \textbf{97.8 $\pm$ 0.4} \\ \hdashline
\multicolumn{1}{l|}{classifier (50-shot)}       &   90.0 $\pm$ 1.4       & \multicolumn{1}{c|}{90.3 $\pm$ 1.4}          &     93.3 $\pm$ 1.1     & \multicolumn{1}{c|}{93.9 $\pm$ 1.2}          &     96.3 $\pm$ 0.6     & \multicolumn{1}{c|}{96.4 $\pm$ 0.6}          &    94.8  $\pm$ 0.7     &  95.1 $\pm$ 0.6        \\ \hline
\multicolumn{1}{l|}{\textbf{10-shot}} & \textbf{Work}                        & \multicolumn{1}{c|}{\textbf{Travel}}                      & \textbf{Work}                        & \multicolumn{1}{c|}{\textbf{Travel}}                      & \textbf{Work}                        & \multicolumn{1}{c|}{\textbf{Travel}}                      & \textbf{Work}                        & \textbf{Travel}                      \\ \hline
\multicolumn{1}{l|}{classifier}       & 86.0 $\pm$ 2.2          & \multicolumn{1}{c|}{93.0 $\pm$ 1.2}          & 97.2 $\pm$ 0.7          & \multicolumn{1}{c|}{94.8 $\pm$ 0.9}          & 94.5 $\pm$ 0.8          & \multicolumn{1}{c|}{96.8 $\pm$ 0.6}          & 95.8 $\pm$ 0.3          & 95.8 $\pm$ 0.4          \\
\multicolumn{1}{l|}{classifier-EDA}   & 86.4 $\pm$ 1.7          & \multicolumn{1}{c|}{93.0 $\pm$ 1.0}          & 97.0 $\pm$ 0.9          & \multicolumn{1}{c|}{95.2 $\pm$ 1.3}          & 94.7 $\pm$ 0.7          & \multicolumn{1}{c|}{96.9 $\pm$ 0.5}          & 95.8 $\pm$ 0.5          & 96.0 $\pm$ 0.6          \\
\multicolumn{1}{l|}{classifier-BT}    & 83.7 $\pm$ 1.7          & \multicolumn{1}{c|}{91.9 $\pm$ 1.1}          & \textbf{97.4 $\pm$ 0.9} & \multicolumn{1}{c|}{\textbf{96.1 $\pm$ 0.8}} & 93.6 $\pm$ 0.7          & \multicolumn{1}{c|}{96.4 $\pm$ 0.5}          & 95.5 $\pm$ 0.6          & \textbf{96.2 $\pm$ 0.3} \\
\multicolumn{1}{l|}{DNNC-scratch}      & 92.6 $\pm$ 1.7          & \multicolumn{1}{c|}{\textbf{96.4 $\pm$ 0.6}} & 94.1 $\pm$ 1.4          & \multicolumn{1}{c|}{84.8 $\pm$ 3.0}          & 98.4 $\pm$ 0.6          & \multicolumn{1}{c|}{\textbf{98.5 $\pm$ 0.3}} & 96.2 $\pm$ 0.8          & 91.1 $\pm$ 1.8          \\
\multicolumn{1}{l|}{DNNC}    & \textbf{95.0 $\pm$ 1.0} & \multicolumn{1}{c|}{96.0 $\pm$ 0.8}          & 95.5 $\pm$ 0.9          & \multicolumn{1}{c|}{93.3 $\pm$ 1.9}          & \textbf{99.0 $\pm$ 0.4} & \multicolumn{1}{c|}{98.3 $\pm$ 0.4}          & \textbf{97.2 $\pm$ 0.5} & 95.7 $\pm$ 1.0          \\ \hdashline
\multicolumn{1}{l|}{classifier (50-shot)}       &  94.3 $\pm$ 0.8      & \multicolumn{1}{c|}{97.0 $\pm$ 0.3}          &   95.2 $\pm$ 0.8       & \multicolumn{1}{c|}{92.6 $\pm$ 1.4}          &     97.7 $\pm$ 0.3     & \multicolumn{1}{c|}{98.6 $\pm$ 0.2}          &     96.5 $\pm$ 0.5     &    95.5 $\pm$ 0.8      \\ \hline
\end{tabular}}\caption{Testing results on the four different domains.}
\label{table:fourDomain}
\vspace{1em}

\tiny
\resizebox{1.0\linewidth}{!}{
\begin{tabular}{l|cl|cl|cl|cc}
\hline
                       & \multicolumn{2}{c|}{\textbf{In-domain accuracy}}                & \multicolumn{2}{c|}{\textbf{OOS recall}}                        & \multicolumn{2}{c|}{\textbf{OOS precision}}                     & \multicolumn{2}{c}{\textbf{OOS F1}}              \\
                       & \textbf{5-shot}         & \multicolumn{1}{c|}{\textbf{10-shot}} & \textbf{5-shot}         & \multicolumn{1}{c|}{\textbf{10-shot}} & \textbf{5-shot}         & \multicolumn{1}{c|}{\textbf{10-shot}} & \textbf{5-shot}                      & \textbf{10-shot} \\ \cline{2-9} 
classifier             & 77.7 $\pm$ 0.6          & 83.2 $\pm$ 0.6                        & \textbf{91.9 $\pm$ 1.1}          & \textbf{92.9 $\pm$ 0.8}               & 53.9 $\pm$ 0.8          & 58.2 $\pm$ 1.0                        & 68.0 $\pm$ 1.0          &      71.6 $\pm$ 0.8            \\
USE+ConveRT             & 76.9 $\pm$ 0.7          &  82.6 $\pm$ 0.5                    & 90.7 $\pm$ 0.8        &    89.1  $\pm$  0.4        & 49.0 $\pm$ 0.8          &       55.1   $\pm$  0.7             & 63.6 $\pm$ 0.7          &   68.1 $\pm$ 0.6          \\
DNNC          & \textbf{84.9 $\pm$ 0.8} & \textbf{91.6 $\pm$ 0.3}               & 90.1 $\pm$ 1.6 & 83.0 $\pm$ 2.0                        & \textbf{64.7 $\pm$ 2.5} & \textbf{81.4 $\pm$ 2.0}               &  \textbf{75.3 $\pm$ 2.0}        &   \textbf{82.1 $\pm$ 0.3}               \\ \hdashline
classifier (full-shot) & \multicolumn{2}{c|}{92.9 $\pm$ 0.2}                & \multicolumn{2}{c|}{90.2 $\pm$ 0.5}                & \multicolumn{2}{c|}{76.7 $\pm$ 0.7}                & \multicolumn{2}{c}{82.9 $\pm$ 0.6}         \\
USE+ConveRT (full-shot) & \multicolumn{2}{c|}{95.0 $\pm$ 0.1}                & \multicolumn{2}{c|}{64.6 $\pm$ 0.6}                & \multicolumn{2}{c|}{79.3 $\pm$ 0.7}                & \multicolumn{2}{c}{71.2 $\pm$ 0.5}         \\ \hline
\end{tabular}}\caption{Testing results on the whole dataset (5 runs).}
\label{table:wholeDataset}
\vspace{-0.5em}

\end{table*}

\paragraph{Using the development set}
We would not always have access to a large enough development set in the few-shot learning scenario.
However, we still use the development set provided by the dataset to investigate the models' behaviors when changing hyper-parameters like the threshold. 

\paragraph{Models compared}
We list the models used in our experiments:

\begin{itemize}
\item{\bf Classifier baselines:}
``Classifier'' is the RoBERTa-based classification model described in Section~\ref{subsec:classifier}.
We further seek solid baselines by data augmentation.
``Classifier-EDA'' is the classifier trained with data augmentation techniques in \citet{eda}.
``Classifier-BT'' is the classifier trained with back-translation data augmentation~\citep{qanet,backtrans-da} by using a transformer-based English$\leftrightarrow$German translation system~\citep{transformer}.

\item{\bf Non-BERT classifier:}
We also test a state-of-the-art fast embedding-based classifier, ``USE+ConveRT''~\citep{convert,efficient_intent}, in the ``all domains'' setting.
\citet{efficient_intent} showed that the ``USE+ConveRT'' outperformed a BERT classifier on the CLINC150 dataset, while it was not evaluated along with the OOS detection task.
We modified their original code\footnote{\url{https://github.com/connorbrinton/polyai-models/releases/tag/v1.0}.} to apply the uncertainty-based OOS detection.

\item{\bf $k$NN baselines:\footnote{We tried weighted voting in \citet{knn-classifier}, but $k=1$ performed better in general.}}
``Emb-$k$NN'' is the $k$NN method with S(Ro)BERT(a) described in Section~\ref{subsec:joint}, and ``Emb-$k$NN-vanilla'' is {\it without} using our intent training examples for fine-tuning.
``TF-IDF-$k$NN'' is another $k$NN baseline using TF-IDF vectors, which tells us how well string matching performs on our task.
We also implement a relation network~\citep{relation_net}, ``RN-$k$NN,'' to learn a similarity metric between the SRoBERTa embeddings, instead of using the cosine similarity.

\item{\bf Proposed method:\footnote{Our code will be available at \url{https://github.com/salesforce/DNNC-few-shot-intent}.}}
``DNNC'' is our proposed method, and ``DNNC-scratch'' is {\it without} the NLI pre-training in Section~\ref{subsec:nli_transfer}.
``DNNC-joint'' is our joint approach on top of top-$k$ retrieval by Emb-$k$NN (Section~\ref{subsec:joint}).

\end{itemize}

\noindent
More details about the model training and the data augmentation configurations are described in  Appendix~\ref{training-details} and Appendix~\ref{data-augmentation}, respectively.

\section{Experimental Results}

This section shows our experimental results.
Appendix~\ref{extra-results} shows some additional figures.

\subsection{Model Performance CLINC150 Dataset}
\paragraph{Single domains}
We first show test set results of 5-shot and 10-shot in-domain classification and OOS detection accuracy in Table~\ref{table:fourDomain} for the four selected domains.
In the 5-shot setting, the proposed DNNC method consistently attains the best results across all the four domains.
The comparison between DNNC-scratch and DNNC shows that our NLI task transfer is effective.
In the 10-shot setting, all the approaches generally experience an accuracy improvement due to the additional training data, and the dominant performance of DNNC weakens, although it remains highly competitive.
We can see that our DNNC is comparable with or even surpasses some of the 50-shot classifier's scores, and the data augmentation techniques are not always helpful when we use the strong pre-trained model.

\paragraph{Entire CLINC150 dataset}
Next, Table~\ref{table:wholeDataset} shows results to compare our method with the classifier and USE+ConveRT baselines, on the entire CLINC150 dataset with the 150 intents. USE+ConveRT performs worse than the RoBERTa-based classifier on the OOD detection task. 
The advantage of DNNC for in-domain intent detection is clear, with its 10-shot in-domain accuracy close to the upper-bound accuracy for the classifier baseline.
One observation is that our DNNC method tends to be more confident about its prediction, with the increasing number of the training examples; as a result, the OOS recall becomes lower in the 10-shot setting, while the OOS precision is much higher than the other baselines.
Better controlling the confidence output of the model is an interesting direction for future work.

When the USE+ConveRT baseline is evaluated along with the OOS detection task, its overall accuracy is not as good as the other RoBERTa-based models, despite its potential in the purely in-domain classification.
This indicates that the fine-tuned (Ro)BERT(a) models are more robust to out-of-distribution examples than shallower models like USE+ConveRT, also suggested in \citet{acl-ood}.
\begin{figure*}[t!]
\centering
\begin{minipage}[t]{1.0\textwidth}
    \centering
    \includegraphics[width=0.92\linewidth]{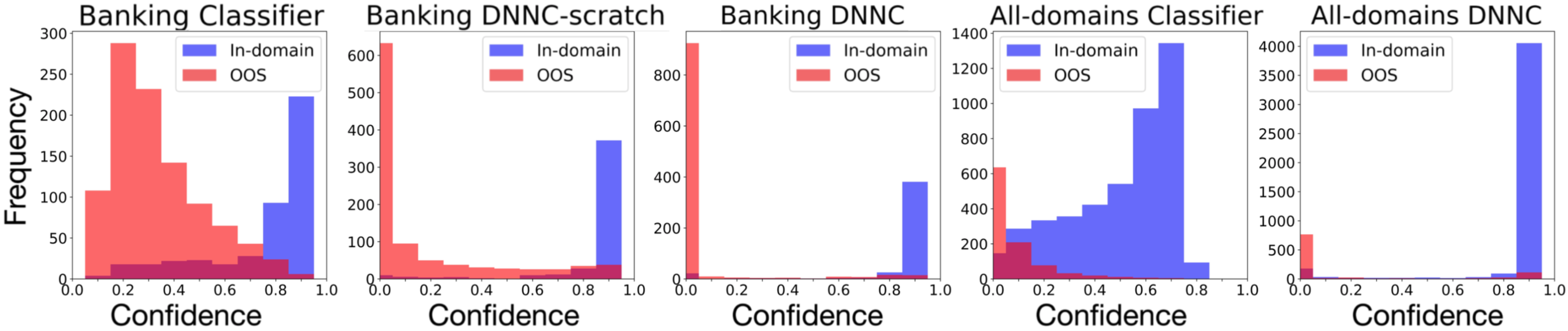}
    \caption{Model confidence level on 5-shot test sets for the banking domain and all domains.}
    \label{fig:separation}
\end{minipage}%

\vspace{10pt}

\begin{minipage}[t]{0.71\textwidth}
    \centering
    \includegraphics[width=0.3\linewidth]{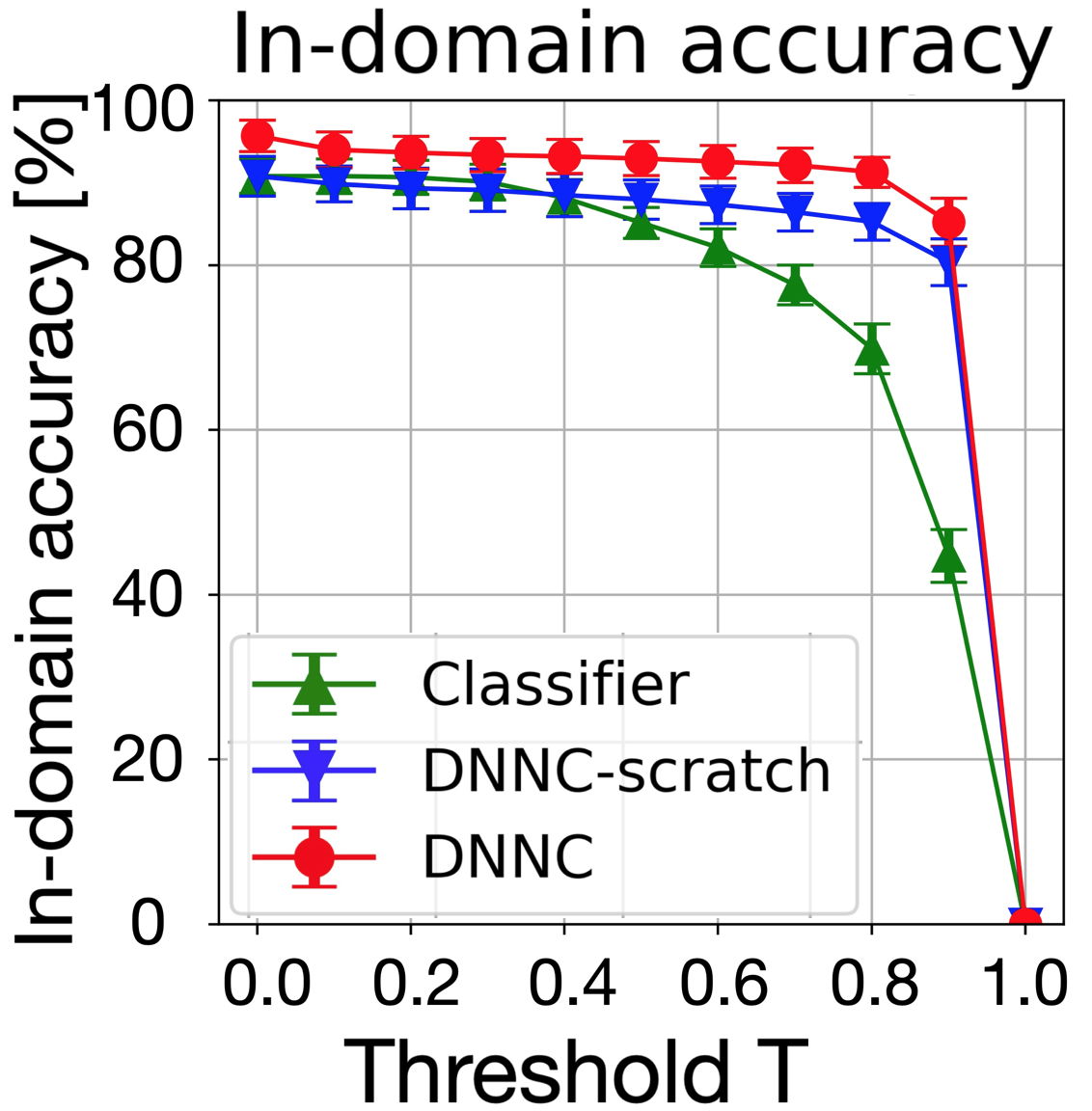}
    \includegraphics[width=0.3\linewidth]{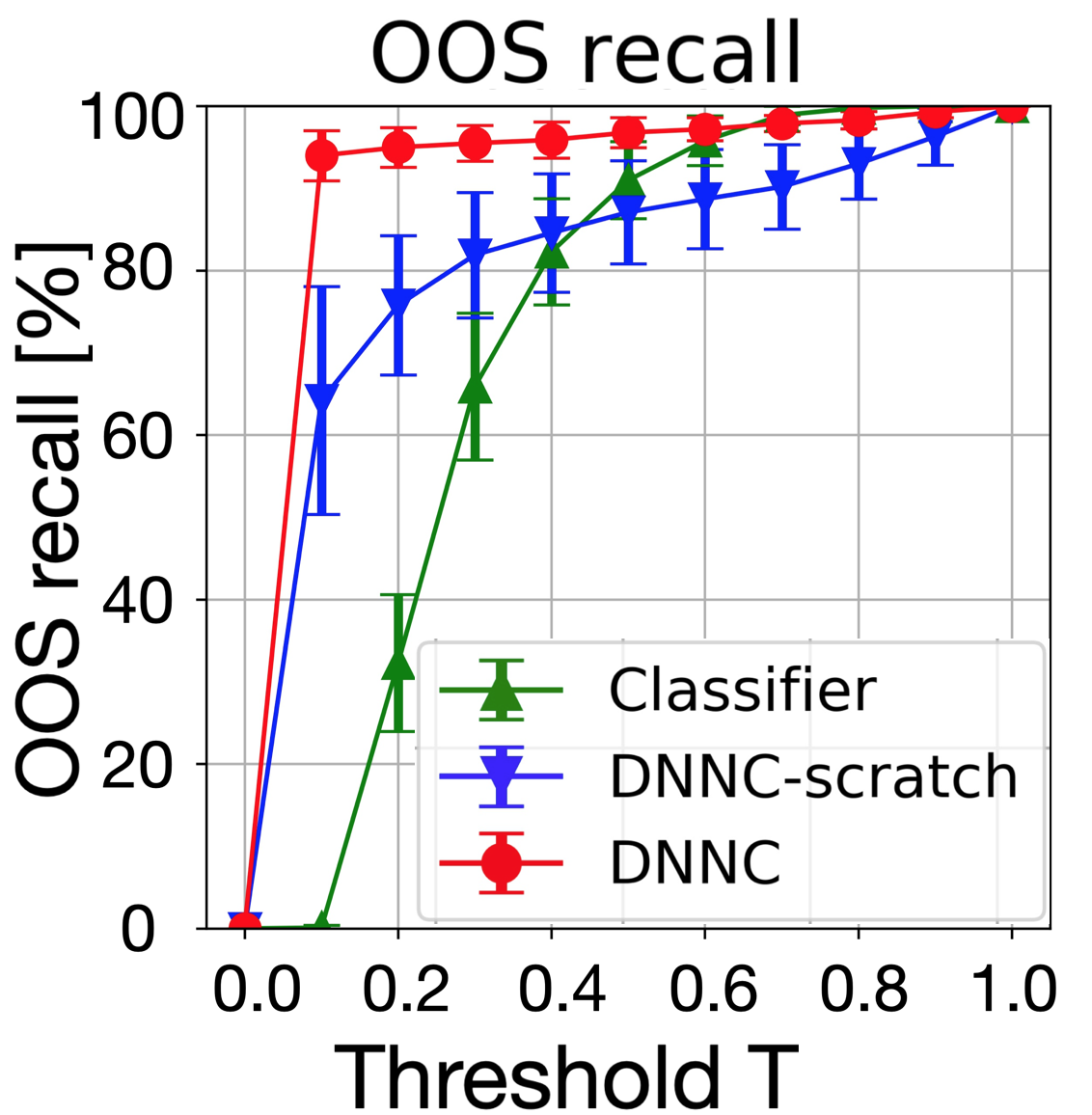}
    \includegraphics[width=0.3\linewidth]{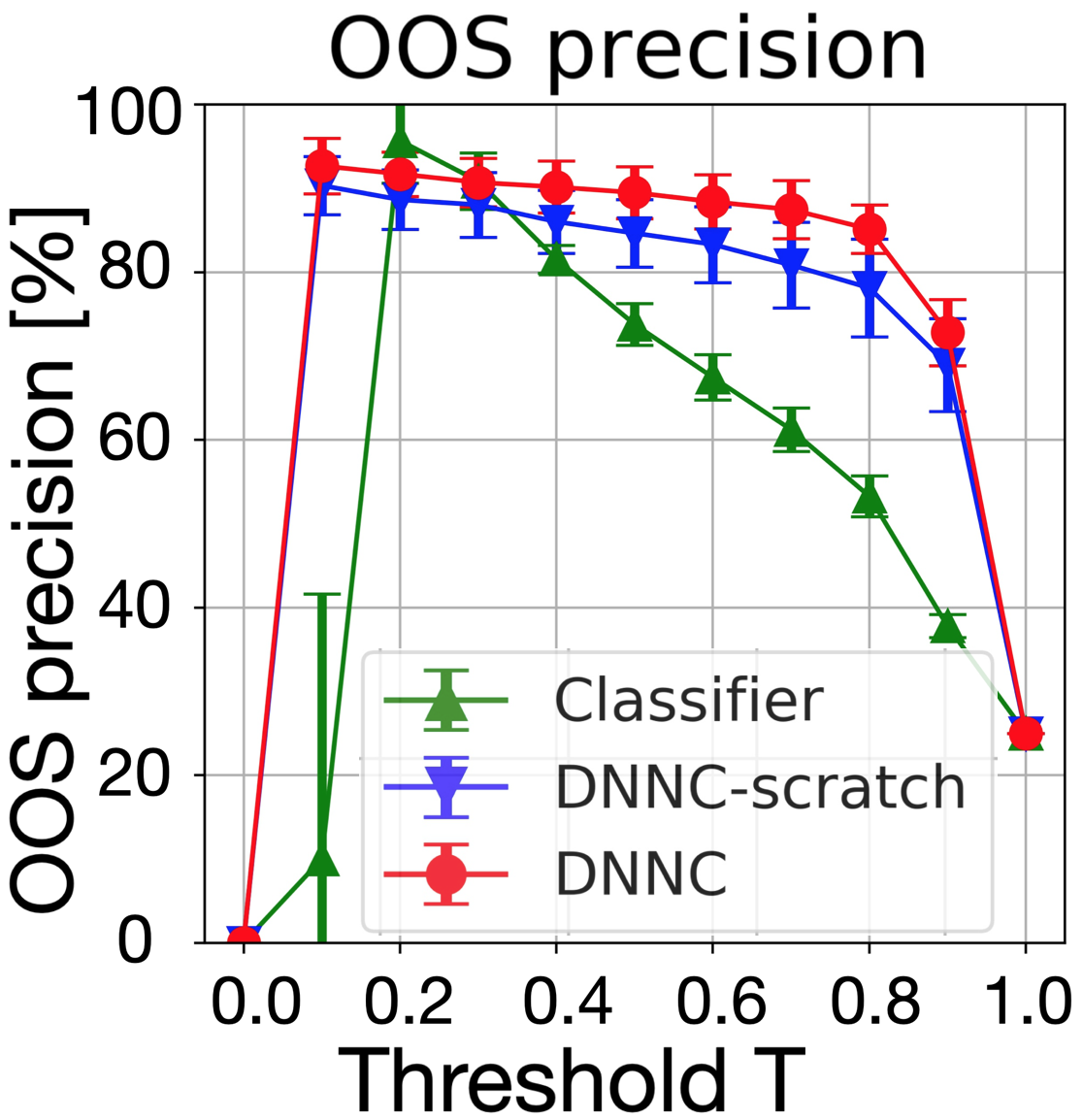}
    \caption{Development set results on the banking domain in the 5-shot setting. In this series of plots, a model with a higher area-under-the-curve is more robust.}
    \label{fig:robust}
    \end{minipage}%
\hfill
\begin{minipage}[t]{0.25\textwidth}
    \centering
    \includegraphics[width=1.\linewidth]{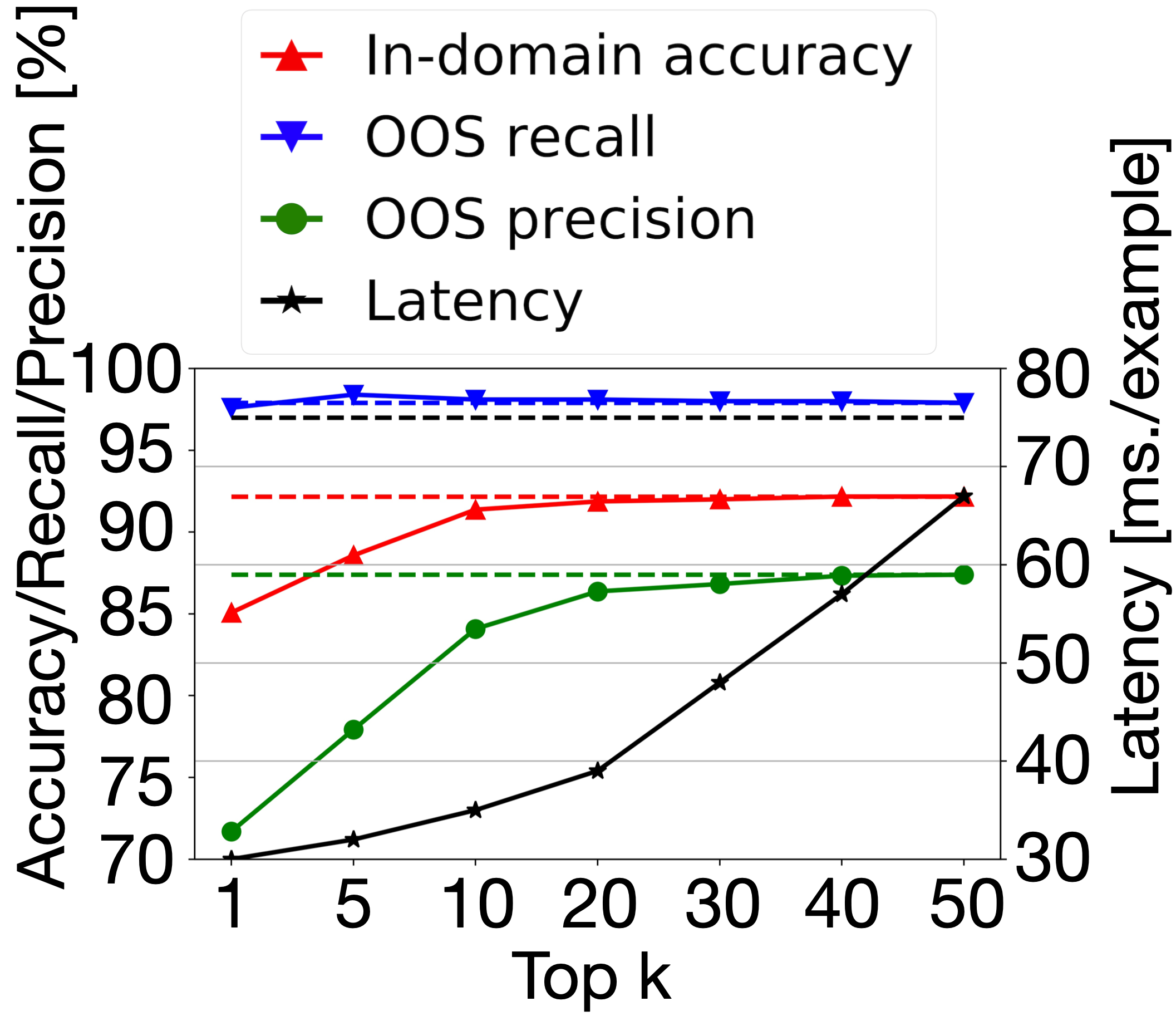}
    \caption{Accuracy vs. latency of DNNC-joint.}
    \label{fig:join-nli}
\end{minipage}%
\centering
\label{fig:compare_fig}
\end{figure*}

\begin{table*}[]
\centering
\small
\resizebox{1.0\linewidth}{!}{
\begin{tabular}{l|cc|cc|cc|cc|rr}
\hline
                     & \multicolumn{2}{c|}{\textbf{In-domain accuracy}}  & \multicolumn{2}{c|}{\textbf{OOS recall}} & \multicolumn{2}{c|}{\textbf{OOS precision}}       & \multicolumn{2}{c}{\textbf{OOS F1}} & \multicolumn{2}{c}{\textbf{Latency [ms./example]}} \\
\textbf{Banking}              & \textbf{5-shot}                & \textbf{10-shot} & \textbf{5-shot}    & \textbf{10-shot}    & \textbf{5-shot}         & \textbf{10-shot}        & \textbf{5-shot}  & \textbf{10-shot} & \textbf{5-shot}   & \textbf{10-shot}  \\ \hline
TF-IDF-$k$NN               & 59.2 $\pm$ 3.6                 & 65.9 $\pm$ 3.8   & 61.0 $\pm$ 2.8     & 51.3 $\pm$ 1.6      & 97.6 $\pm$ 0.6 & 98.2 $\pm$ 0.3 &      75.1  $\pm$ 2.2           &    67.4 $\pm$ 1.3              & 1                 & 1                 \\
Emb-$k$NN-vanilla           &   64.4 $\pm$ 2.8              &  65.4 $\pm$ 2.3    &  89.5 $\pm$ 1.1   &  94.6 $\pm$ 0.5     &     95.1 $\pm$ 0.5       &  92.6 $\pm$ 0.7       & 92.2 $\pm$ 0.5  & 93.6 $\pm$ 0.5  &     15        & 15           \\
Emb-$k$NN            & 78.4 $\pm$ 2.4                 & 84.3 $\pm$ 1.2   & 92.0 $\pm$ 2.0     & 91.6 $\pm$ 1.2      & 91.4 $\pm$ 0.8          & 93.7 $\pm$ 0.5          & 91.7 $\pm$ 0.7   & 92.6 $\pm$ 0.7   & 15             & 15             \\
RN-$k$NN            &     79.5  $\pm$ 3.1           & 89.0 $\pm$ 1.4    &  88.3 $\pm$ 1.9    &  75.9 $\pm$ 4.0     &    91.6 $\pm$ 1.3       &   95.9 $\pm$ 0.6     &  89.9 $\pm$ 1.0  & 84.7 $\pm$ 2.5 & 17             & 17             \\ \hdashline
DNNC-joint   & 88.5 $\pm$ 1.2                 & 91.0 $\pm$ 1.0   & 95.0 $\pm$ 0.9     & 95.2 $\pm$ 1.1      & 96.9 $\pm$ 0.4          & 97.3 $\pm$ 0.4          & 96.0 $\pm$ 0.5   & 96.3 $\pm$ 0.6   & 36             & 36             \\ 
DNNC        & 88.6 $\pm$ 1.3                 & 91.2 $\pm$ 1.1   & 94.7 $\pm$ 1.0     & 94.8 $\pm$ 1.1      & 97.0 $\pm$ 0.3          & 97.5 $\pm$ 0.4          & 95.9 $\pm$ 0.5   & 96.1 $\pm$ 0.6   & 73             & 143             \\ \hline
\textbf{All domains} & \textbf{5-shot}                & \textbf{10-shot} & \textbf{5-shot}    & \textbf{10-shot}    & \textbf{5-shot}         & \textbf{10-shot}        & \textbf{5-shot}  & \textbf{10-shot} & \textbf{5-shot}   & \textbf{10-shot}  \\ \hline
TF-IDF-$k$NN               & 35.4 $\pm$ 0.7                 & 31.3 $\pm$ 1.1   & 72.2 $\pm$ 2.1     & 83.1 $\pm$ 1.5      & 24.7 $\pm$ 0.4          & 23.5 $\pm$ 0.4          &     36.8 $\pm$    0.6          &   36.6 $\pm$ 0.6              & 1             & 2             \\
Emb-$k$NN-vanilla           &  54.2 $\pm$ 0.5              &  50.7 $\pm$ 0.9 &   87.5 $\pm$ 0.7  &   95.1 $\pm$ 0.3  &  39.6 $\pm$ 0.3       &  34.2 $\pm$ 0.3    & 54.5 $\pm$ 0.4 & 50.3 $\pm$ 0.3  &   16        &     19         \\
Emb-$k$NN            &         78.7 $\pm$ 1.0                       &     86.9  $\pm$ 0.3            &     86.8 $\pm$ 2.4               &   82.3 $\pm$ 0.8                 &           55.0 $\pm$ 1.5              &      66.5 $\pm$ 1.0                   &        67.3 $\pm$ 1.6          &       73.5 $\pm$ 0.7          &         16          &     19              \\
RN-$k$NN             &    76.6 $\pm$ 0.6      &      88.7  $\pm$ 1.1        &    88.9 $\pm$ 1.4                &     76.6 $\pm$ 2.7               &   50.0 $\pm$ 0.3                       &      71.0 $\pm$ 0.5                  &           64.0   $\pm$ 0.2     &       73.7 $\pm$ 1.0         &     16              &    18               \\ \hdashline
DNNC-joint   & 84.5 $\pm$ 0.8 & 91.2 $\pm$ 0.2   & 90.6 $\pm$ 1.6     & 85.1 $\pm$ 1.8      & 63.6 $\pm$ 2.4          & 78.4 $\pm$ 2.1          &        74.7  $\pm$ 1.9         &     81.6 $\pm$ 0.4             & 37             & 41             \\ 
DNNC        & 84.9 $\pm$ 0.8                 & 91.6 $\pm$ 0.3   & 90.1 $\pm$ 1.6     & 83.0 $\pm$ 2.0      & 64.7 $\pm$ 2.5          & 81.4 $\pm$ 2.0          &            75.3 $\pm$ 2.0       &   82.1 $\pm$ 0.3                        & 697             & 1498             \\ \hline
\end{tabular}}\caption{Comparison among the nearest neighbor methods on the test sets for the banking domain and all the domains. The latency is measured on a single NVIDIA Tesla V100 GPU, where the batch size is $1$ to simulate an online use case. DNNC-joint is based on top-20 Emb-$k$NN retrieval. }
\label{table:latency}
\end{table*}

\subsection{Robustness of DNNC}
As described in Section~\ref{subsection:evalMetrics}, we select the threshold to determine OOS by making a trade-off between in-domain classification and OOS detection accuracy.
It is therefore desirable to have a model with candidate thresholds that provide high in-domain accuracy as well as OOS precision and recall. 

We observe in Figure~\ref{fig:robust} that in the 5-shot setting, DNNC is the most robust to the threshold selection.
The contrast between the classification model and DNNC-scratch suggests that nearest neighbor approaches (in this case DNNC) make for stronger discriminators; the advantage of DNNC over DNNC-scratch further demonstrates the power of the NLI transfer and, perhaps more importantly, the effectiveness of the pairwise discriminative pre-training.
This result is consistent with the intuition we gained from Figure~\ref{fig:tsne-main}, and the overall observation is also consistent across different settings.


To further understand the differences in behaviors between the classification model and DNNC method, we examine the output from the final softmax/sigmoid function (model confidence score) in Figure~\ref{fig:separation}.
At 5-shot, the classifier method still struggles to fully distinguish the in-domain examples from the OOS examples in its confidence scoring, while DNNC already attains a clear distinction between the two.
Again, we can clearly see the effectiveness of the NLI transfer.

With the model architectures for BERT-based classifier and DNNC being the same (RoBERTa is used for both methods) except for the final layer (multi-class-softmax vs. binary sigmoid), this result suggests that the pairwise NLI-like training is more sample-efficient, making it an excellent candidate for the few-shot use case.


\subsection{DNNC-joint for Faster Inference}
Despite its effectiveness in few-shot intent and OOS settings, the proposed DNNC method might not scale in high-traffic use cases, especially when the number of classes, $N$, is large, due to the inference-time bottleneck (Section~\ref{subsec:joint}).
With this in mind, we proposed the DNNC-joint approach, wherein a faster model is used to filter candidates for the fine-tuned DNNC model. 

We compare the accuracy and inference latency metrics for various methods in Table~\ref{table:latency}.
Note that Emb-$k$NN and RN-$k$NN exhibit excellent latency performance, but they fall considerably short in both the in-domain intent and OOS detection accuracy, compared to DNNC and the DNNC-joint methods.
On the other hand, the DNNC-joint model shows competitiveness in both inference latency and accuracy.
These results indicate that the current text embedding approaches like SBERT are not enough to fully capture fine-grained semantics.

Intuitively, there is a trade-off between latency and inference accuracy: with aggressive filtering, the DNNC inference step needs to handle a smaller number of training examples, but might miss informative examples; with less aggressive filtering, the NLI model sees more training examples during inference, but will take longer to process single user input.
This is illustrated in Figure~\ref{fig:join-nli}, where the in-domain intent and OOS accuracy metrics (on the development set of the banking domain in the 5-shot setting) improve with the increase of $k$, while the latency increases at the same time.
Empirically, $k=20$ appears to strike the balance between latency and accuracy, with the accuracy metrics similar to those of the DNNC method, while being much faster than DNNC (dashed lines are the corresponding DNNC references).

\section{Discussions and Related Work}

\paragraph{Interpretability}
Interpretability is an important line of research recently~\citep{interpret-qa-1,interpret-qa-2,graph-retriever}.
The nearest neighbor approach~\citep{knn-nips} is appealing in that we can explicitly know which training example triggers each prediction.
Table~\ref{table:case-study} in Appendix~\ref{extra-results} shows some examples.

\paragraph{Call for better embeddings}
Emb-$k$NN and RN-$k$NN are not as competitive as DNNC.
This encourages future work on the task-oriented evaluation of text embeddings in $k$NN.

\paragraph{Training time}
Our DNNC method needs longer training time than that of the classifier (e.g., 90 vs. 40 seconds to train a single-domain model), because we synthesize the pairwise examples.
As a first step, we used all the training examples to investigate the effectiveness, but it is an interesting direction to seek more efficient pairwise training.

\paragraph{Distilled model}
Another way to speedup our model is to use distilled pre-trained models~\citep{distil-bert}.
We replaced the RoBERTa model with a distilled RoBERTa model,
and observed large variances with significantly lower OOS accuracy.
\citet{acl-ood} also suggested that the distilled models would not be robust to out-of-distribution examples.

\paragraph{Few-shot text classification}
Few-shot classification~\citep{fei2006one,vinyals2016matching} has been applied to text classification tasks~\citep{deng2019low,geng2019induction,xu2019open}, and few-shot intent detection is also studied but without OOS~\citep{marrying-reg,xia2020cg,efficient_intent}.
There are two common scenarios: 1) learning with plenty of examples and then generalizing to unseen classes with a few examples, and 2) learning with a few examples for all seen classes.
Meta-learning~\citep{finn2017model,geng2019induction} is widely studied in the first scenario.
In our paper, we have focused on the second scenario, assuming that there are only a limited number of training examples for each class.
Our work is related to metric-based approaches such as matching networks~\cite{matching-net}, prototypical networks~\cite{proto-net} and relation networks~\cite{relation_net}, as they model nearest neighbours in an example-embedding or a class-embedding space.
We showed that a relation network with the RoBERTa embeddings does not perform comparably to our method.
We also considered several ideas from prototypical networks~\citep{hierarchical-proto-net}, but those did not outperform our Emb-$k$NN baseline.
These results indicate that deep self-attention is the key to the nearest neighbor approach with OOS detection.

\section{Conclusion}
In this paper, we have presented a simple yet efficient nearest-neighbor classification model to detect user intents and OOS intents.
It includes paired encoding and discriminative training to model relations between the input and example utterances.
Moreover, a seamless transfer from NLI and a joint approach with fast retrieval are designed to improve the performance in terms of the accuracy and inference speed.
Experimental results show superior performance of our method on a large-scale multi-domain intent detection dataset with OOS.
Future work includes its cross-lingual transfer and cross-dataset (or cross-task) generalization.

\section*{Acknowledgments}
This work is supported in part by NSF under grants III-1763325, III-1909323, and SaTC-1930941.
We thank Huan Wang, Wenpeng Yin for their insightful discussions, and the anonymous reviewers for their helpful and thoughtful comments.
We also thank Jin Qu, Tian Xie, Xinyi Yang, and Yingbo Zhou for their support in the deployment of DNNC into the internal system.


\bibliographystyle{acl_natbib}
\bibliography{emnlp2020}

\appendix

\section*{Appendix}

\begin{table*}[h]
\centering
\begin{tabular}{l|lll}
\hline
Dataset                    & SNLI & WNLI & MNLI \\ \hline
Size of the development set & 9999          & 70            & 9814          \\
Accuracy                   & 94.5\%        & 41.4\%        & 92.1\%        \\ \hline
\end{tabular}\caption{Development results on three NLI datasets.}
\label{table-pretrain}
\end{table*}

\section{Training Details} \label{training-details}

\paragraph{Dataset preparation}
To use the CLINC150 dataset~\citep{oos-intent}\footnote{\url{https://github.com/clinc/oos-eval}.} in our ways, especially for the single-domain experiments, we provide preprocessing scrips accompanied with our code.

\paragraph{General training}\label{appendix-general-training}
This section describes the details about the model training in Section~\ref{subsec:model_setting}.
For each component related to RoBERTa and SRoBERTa, we solely follow the two libraries, transformers and sentence-transformers, for the sake of easy reproduction of our experiments.\footnote{\url{https://github.com/huggingface/transformers} and \url{https://github.com/UKPLab/sentence-transformers}.}
The example code to train the NLI-style models is also available.\footnote{\url{https://github.com/huggingface/transformers/tree/master/examples/text-classification}.}
We use the {\tt roberta-base} configuration\footnote{\url{https://s3.amazonaws.com/models.huggingface.co/bert/roberta-base-config.json}.} for all the RoBERTa/SRoBERTa-based models in our experiments.
All the model parameters including the RoBERTa parameters are updated during all the fine-tuning processes, where we use the AdamW~\citep{adamw} optimizer with a weight decay coefficient of 0.01 for all the non-bias parameters.
We use a gradient clipping technique~\citep{clip} with a clipping value of 1.0, and also use a linear warmup learning-rate scheduling with a proportion of 0.1 with respect to the maximum number of training epochs.

\begin{table*}[t]
\centering
\resizebox{1.0\linewidth}{!}{
\begin{tabular}{l|ccc|ccc}
\hline
           & \multicolumn{3}{c|}{\textbf{Single domain}}                 & \multicolumn{3}{c}{\textbf{All domains}}                 \\ \cline{2-7} 
           & \textbf{Learning rate} & \textbf{Epoch}    & \textbf{Run} & \textbf{Learning rate} & \textbf{Epoch} & \textbf{Run} \\ \hline
Classifier & \{1e-4, 2e-5, 5e-5\}   & \{15, 25, 35\}    & 10             & \{1e-4, 5e-5\}         & \{15, 25, 35\} & 5              \\
Emb-kNN    & \{1e-4, 2e-5, 3e-5\}   & \{7, 10, 20, 25, 35\} & 10             &   \{2e-5, 5e-5\}        &  \{3, 5, 7\}        & 5              \\
DNNC       & \{1e-5, 2e-5, 3e-5, 4e-5\}   & \{7, 10, 15\}     & 10             & \{2e-5, 5e-5\}         & \{3, 5, 7\}   & 5              \\ \hline
\end{tabular}}\caption{Some hyper-parameter settings for a few models.}\label{table:hyper-paramter}
\end{table*}

\begin{table*}[t]
\centering
\begin{tabular}{l|cc}
\hline
           & \textbf{5-shot}                 & \textbf{10-shot}                \\ \hline
Classifier & \{bs: 50, ep: 25.0, lr: 5e-05\} & \{bs: 50, ep: 35.0, lr: 5e-05\} \\ \hline
Emb-kNN    & \{bs: 200, ep: 7.0, lr: 2e-05\} & \{bs: 200, ep: 5.0, lr: 2e-05\}  \\ \hline
DNNC       & \{bs: 900, ep: 7.0, lr: 2e-05\} &  \{bs: 1800, ep: 5.0, lr: 2e-05\} \\ \hline
\end{tabular}
\caption{Best hyper-parameter settings for a few models on the all-domain experiments, where {\tt bs} is batch size, {\tt ep} represents epochs, {\tt lr} is learning rate.}\label{table:hyper-paramter-best-all}
\end{table*}

\begin{table*}[t]
\centering
\resizebox{1.0\linewidth}{!}{
\begin{tabular}{l|cccc}
\hline
           & \textbf{5-shot}                                      & \multicolumn{1}{c|}{\textbf{10-shot}}                 & \textbf{5-shot}                  & \textbf{10-shot}                 \\ \cline{2-5} 
           & \multicolumn{2}{c|}{\textbf{Banking}}                                                                                 & \multicolumn{2}{c}{\textbf{Credit cards}}                                    \\ \hline
Classifier & \multicolumn{1}{l}{\{bs: 15, ep: 25.0, lr: 5e-05\}}  & \multicolumn{1}{l|}{\{bs: 15, ep: 35.0, lr: 5e-05\}}  & \{bs: 15, ep: 15.0, lr: 5e-05\}  & \{bs: 15, ep: 25.0, lr: 5e-05\}  \\
Emb-kNN    & \multicolumn{1}{l}{\{bs: 200, ep: 35.0, lr: 1e-05\}} & \multicolumn{1}{l|}{\{bs: 200, ep: 25.0, lr: 2e-05\}} & \{bs: 100, ep: 20.0, lr: 1e-05\} & \{bs: 100, ep: 10.0, lr: 1e-05\} \\
DNNC       & \multicolumn{1}{l}{\{bs: 370, ep: 15.0, lr: 1e-05\}} & \multicolumn{1}{l|}{\{bs: 370, ep: 7.0, lr: 2e-05\}}  & \{bs: 370, ep: 15.0, lr: 2e-05\} & \{bs: 370, ep: 7.0, lr: 3e-05\}  \\ \hline
           & \multicolumn{2}{c}{\textbf{Work}}                                                                            & \multicolumn{2}{c}{\textbf{Travel}}                                 \\ \hline
Classifier & \{bs: 15, ep: 15.0, lr: 5e-05\}                      & \{bs: 15, ep: 15.0, lr: 5e-05\}                       & \{bs: 15, ep: 35.0, lr: 5e-05\}  & \{bs: 15, ep: 25.0, lr: 1e-04\}  \\
Emb-kNN    & \{bs: 100, ep: 20.0, lr: 1e-05\}                     & \{bs: 100, ep: 7.0, lr: 2e-05\}                       & \{bs: 100, ep: 35.0, lr: 3e-05\} & \{bs: 100, ep: 20.0, lr: 1e-05\} \\
DNNC       & \{bs: 370, ep: 7.0, lr: 3e-05\}                      & \{bs: 370, ep: 15.0, lr: 2e-05\}                      & \{bs: 370, ep: 7.0, lr: 2e-05\}  & \{bs: 370, ep: 7.0, lr: 2e-05\}  \\ \hline
\end{tabular}}\caption{Best hyper-parameter settings for a few models on the four single domains, where {\tt bs} is batch size, {\tt ep} represents epochs, {\tt lr} is learning rate.}\label{table:hyper-paramter-best}
\end{table*}

\begin{table*}[t]
  \begin{center}
{\small
\resizebox{1.0\linewidth}{!}{
    \begin{tabular}{l|l|l}
    
    Original utterance & Augmented example & Intent label \\ \hline
    can you block my chase account right away please & can you turn my chase account off directly & freeze account \\
    do a car payment from my savings account & with my saving account, you can pay a car payment account & pay bill \\
    when is my visa due & when is my visa to be paid & bill due \\ \hline

    \end{tabular}}
}
    \caption{Examples used to train clasifier-BT.}
    \label{tb:bt-examples}
  \end{center}

\end{table*}

\paragraph{Pre-training on NLI tasks}\label{appendix-pre-training}
For the pre-training on NLI tasks, we fine-tune a {\tt roberta-base} model on three publicly available datasets, i.e., SNLI~\citep{snli}, MNLI~\citep{mnli}, and WNLI~\citep{wnli} from the GLUE benchmark~\citep{glue}.
The optimizer and gradient clipping follow the above configurations.
The number of training epochs is set to $4$; the batch size is set to $32$; the learning rate is set to $2e-5$.
We use a linear warmup learning-rate scheduling with a proportion of $0.06$ by following \citet{roberta}.
The evaluation results on the development sets are shown in Table~\ref{table-pretrain}, where the low accuracy of WNLI is mainly caused by the data size imbalance.
We note that these NLI scores are not comparable with existing NLI scores, because we converted the task to the binary classification task for our model transfer purpose.

\paragraph{Text pre-processing}
For all the RoBERTa-based models, we used the RoBERTa {\tt roberta-base}'s tokenizer provided in the transformers library.\footnote{\url{https://github.com/huggingface/transformers/blob/master/src/transformers/tokenization_roberta.py}.}
We did not perform any additional pre-processing in our experiments.

\paragraph{Hyper-parameter settings}\label{appendix-hyper-parameter}
Table~\ref{table:hyper-paramter} shows the hyper-parameters we tuned on the development sets in our RoBERTa-based experiments.
For a single-domain experiment, we take a hyper-parameter set and apply it to the ten different runs to select the threshold in Section~\ref{subsection:evalMetrics} on the development set.
We then select the best hyper-parameter set along with the corresponding threshold, which achieves the best $J_\mathrm{in\_oos}$ in Equation~(\ref{eq:joint_score}) on the development set, among all the possible hyper-parameter sets.
Finally, we apply the selected model and the threshold to the test set.
We follow the same process for the all-domain experiments, except that we run each experiment five times.
Table~\ref{table:hyper-paramter-best-all} and Table~\ref{table:hyper-paramter-best} summarize the hyper-parameter settings used for the evaluation on the test sets.
We note that each model was not very sensitive to the different hyper-parameter settings, as long as we have a large number of training iterations.

\section{Data Augmentation} \label{data-augmentation}

We describe the details about the classifier baselines with the data augmentation techniques in Section~\ref{subsec:model_setting}.

\paragraph{EDA}
Classifier-EDA uses the following four data augmentation techniques in \citet{eda}: synonym replacement, random insertion, random swap, and random deletion.
We follow the publicly available code.\footnote{\url{https://github.com/jasonwei20/eda_nlp}.}
For every training example, we empirically set one augmentation based on every technique.
We apply each technique separately to the original sentence and therefore every training example will have four augmentations.
The probability of a word in an utterance being edited is set to 0.1 for all the techniques.   

\paragraph{BT}
For classifier-BT, we use the English-German corpus in \citet{escape}, which is widely used in an annual competition for automatic post-editing research on IT-domain text~\citep{ape-2019}.
The corpus contains about 7.5 million translation pairs, and we follow the {\it base} configuration to train a transformer model~\citep{transformer} for each direction.
Based on the initial trial in our preliminary experiments to generate diverse examples, we decided to use a temperature sampling technique instead of a greedy or beam-search strategy.
More specifically, logit vectors during the machine translation process are multiplied by $\tau$ to distort the output distributions, where we set $\tau = 5.0$.
For each training example in the intent detection dataset, we first translate it into German and then translate it back to English.
We repeat this process to generate up to five unique examples, and use them to train the classifier model.
Table~\ref{tb:bt-examples} shows such examples, and we will release all the augmented examples for future research.

\begin{table*}[]
\centering
\resizebox{1.0\linewidth}{!}{
\begin{tabular}{ll}
\hline
\textbf{input utterance}    & transfer ten dollars from my wells fargo account to my bank of america account              \\
\textbf{matched utterance} & transfer \$10 from checking to savings                                                      \\
\textbf{label of the input utterance}      & transfer                                                                                    \\
\textbf{label of the matched utterance}      & transfer                                                                                    \\
\textbf{confidence score}             & 0.934                                                                                      \\ \hline
\textbf{input utterance}    & what transactions have i accrued buying dog food                                            \\
\textbf{matched utterance} & what have i spent on food recently                                                          \\
\textbf{label of the input utterance}      & transactions                                                                                \\
\textbf{label of the matched utterance}      & spending history                                                                           \\
\textbf{confidence score}             & 0.915                                                                                      \\ \hline
\textbf{input utterance}    & who has the best record in the nfl                                                          \\
\textbf{matched utterance} & do i have enough in my boa account for a new pair of skis                                   \\
\textbf{label of the input utterance}      & OOS                                                                                         \\
\textbf{label of the matched utterance}      & balance                                                                                     \\
\textbf{confidence score}             & 0.006                                                                                       \\ \hline
\textbf{input utterance}    & how long will it take me to pay off my card if i pay an extra \$50 a month over the minimum \\
\textbf{matched utterance} & how long do i have left to pay for my chase credit card                                     \\
\textbf{label of the input utterance}      & OOS                                                                                         \\
\textbf{label of the matched utterance}      & bill due                                                                                   \\
\textbf{confidence score}             & 0.945                                                                                      \\ \hline
\end{tabular}} \caption{Case studies on the development set of banking domain. The first two cases are in-domain examples from the banking domain, and the rest are OOS examples.}\label{table:case-study}
\end{table*}

\section{Additional Results}
\label{extra-results}

\begin{figure*}[t]
	\begin{center}
    	\includegraphics[width=0.85\linewidth]{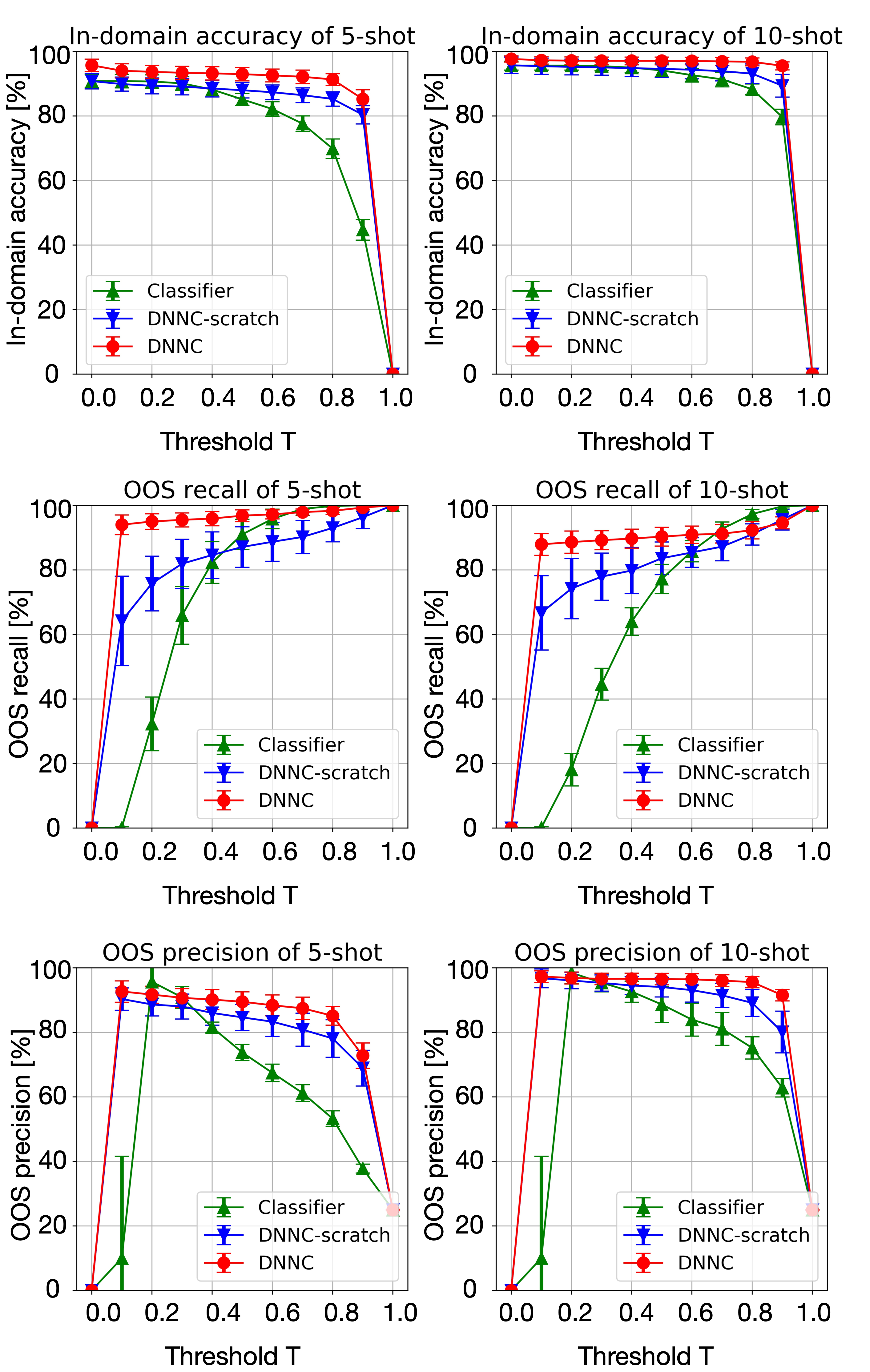}
    \end{center}
\caption{5-shot and 10-shot development results on the banking domain. In this series of plots, a model with a higher area-under-the-curve is more robust.}
\label{fig:visulization-appendix}
\end{figure*}

\begin{figure*}[t]
	\begin{center}
    	\includegraphics[width=0.85\linewidth,height=0.9\textheight]{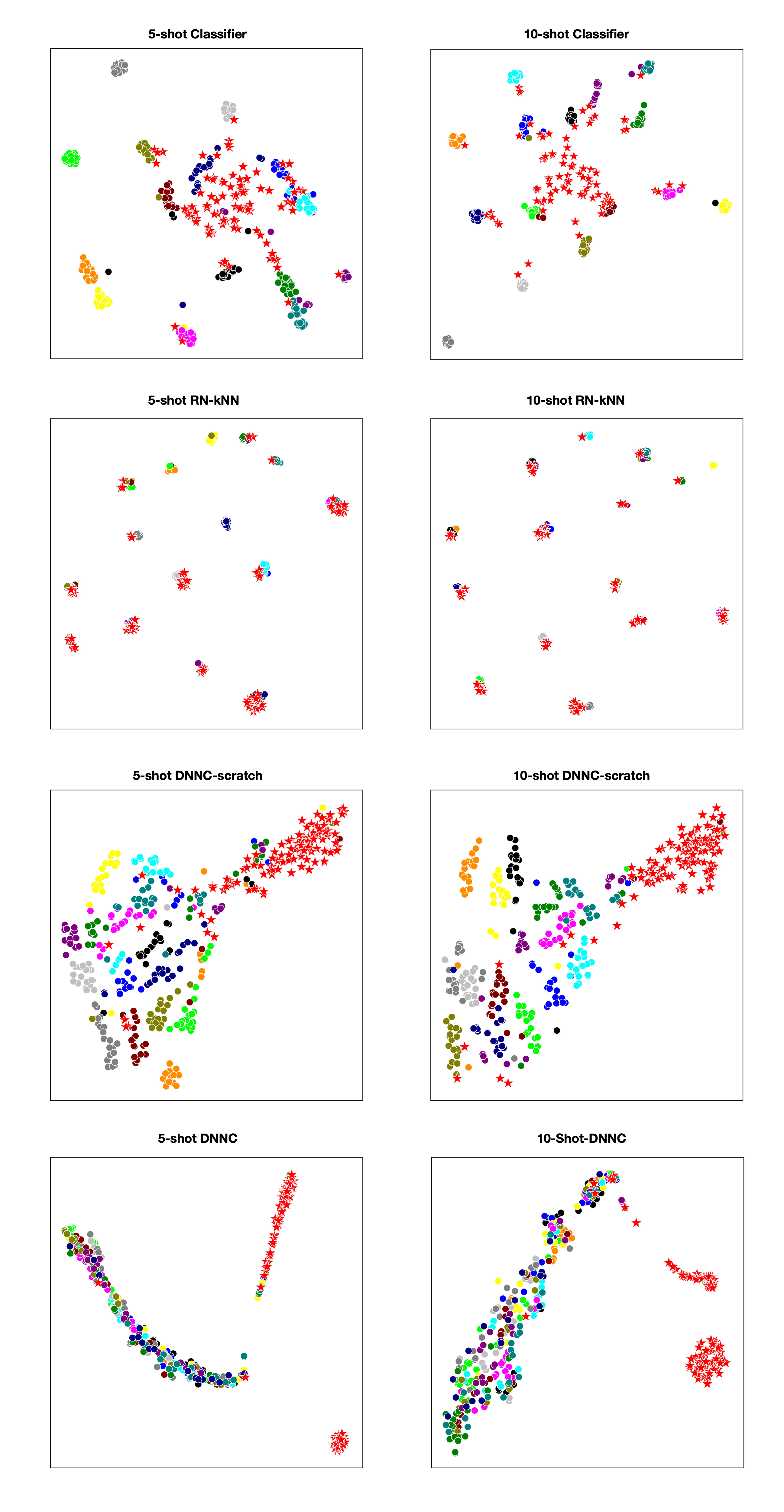}
    \end{center}
\caption{5-shot and 10-shot tSNE visualizations on development set of the banking domain, where circles represent in-domain intent classes, and red stars represent out-of-scope intents.}
\label{fig:tsne-appendix}
\end{figure*}

\paragraph{Visualization} \label{appendix-vidualization}
Figure~\ref{fig:visulization-appendix} shows the same curves in Figure~\ref{fig:robust} along with the corresponding 10-shot results.
We can see that the 10-shot results also exhibit the same trend.
Figure~\ref{fig:tsne-appendix} shows more visualization results with respect to Figure~\ref{fig:tsne-main}.
Again, the 10-shot visualization shows the same trend.

\begin{figure*}[t]
	\begin{center}
    	\includegraphics[width=0.85\linewidth,keepaspectratio=true,height=0.95\textheight]{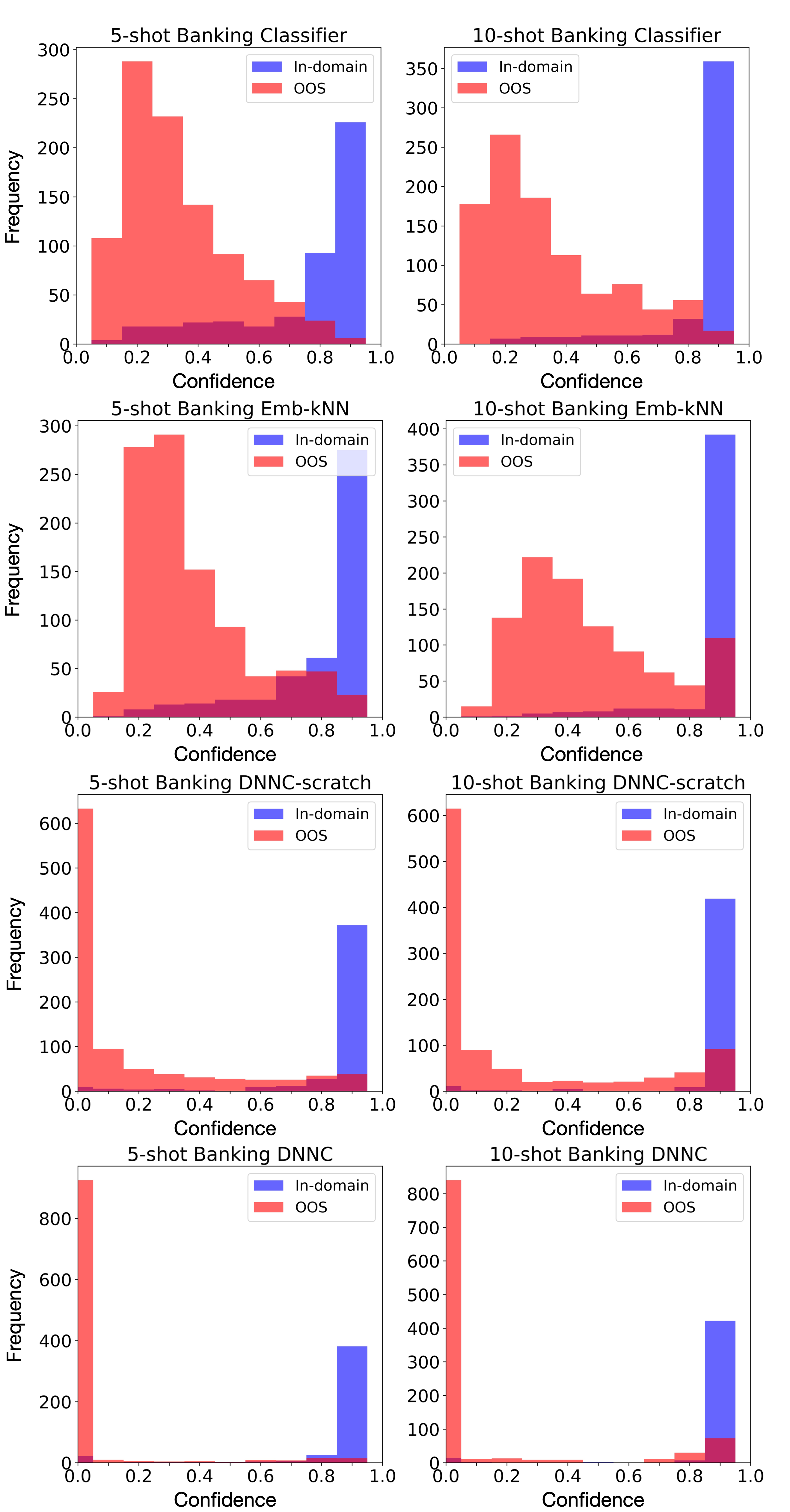}
    \end{center}
\caption{5-shot and 10-shot confidence levels on test set of the banking domain. Best viewed in color.}
\label{fig:Conf-appendix}
\end{figure*}

\begin{figure*}[t]
	\begin{center}
    	\includegraphics[width=0.85\linewidth]{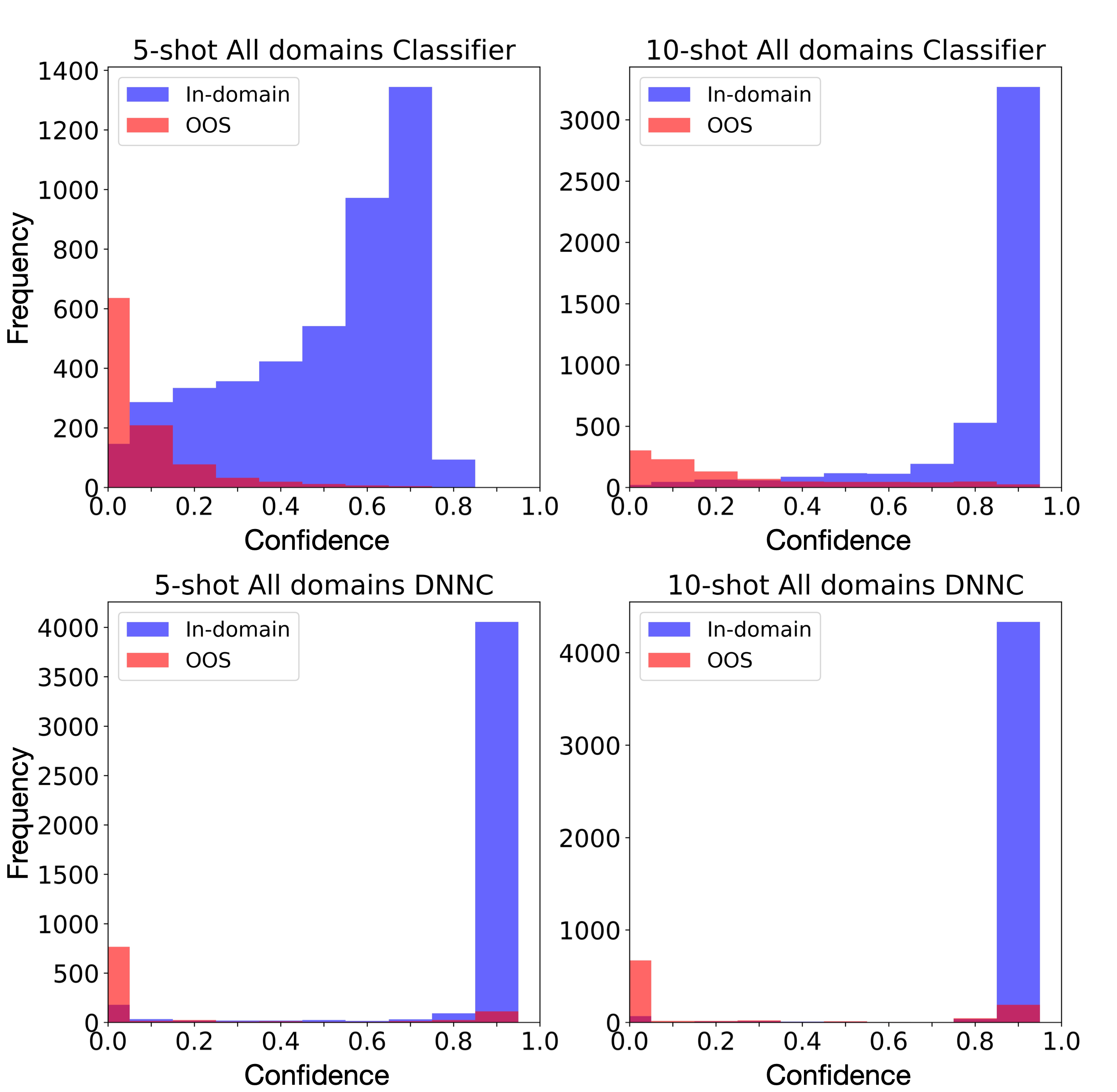}
    \end{center}
\caption{5-shot and 10-shot confidence levels on test set of all domains. Best viewed in color.}
\label{fig:Conf-appendix-all-domains}
\end{figure*}

Figure~\ref{fig:Conf-appendix} and Figure \ref{fig:Conf-appendix-all-domains} show 5-shot and 10-shot confidence levels on the test sets of the banking domain and all domains, respectively.
Both Classifier and Emb-kNN cannot perform well to distinguish the in-domain examples from the OOS examples, while DNNC has a clearer distinction between the two.

\begin{figure*}[t]
	\begin{center}
    	\includegraphics[width=0.85\linewidth]{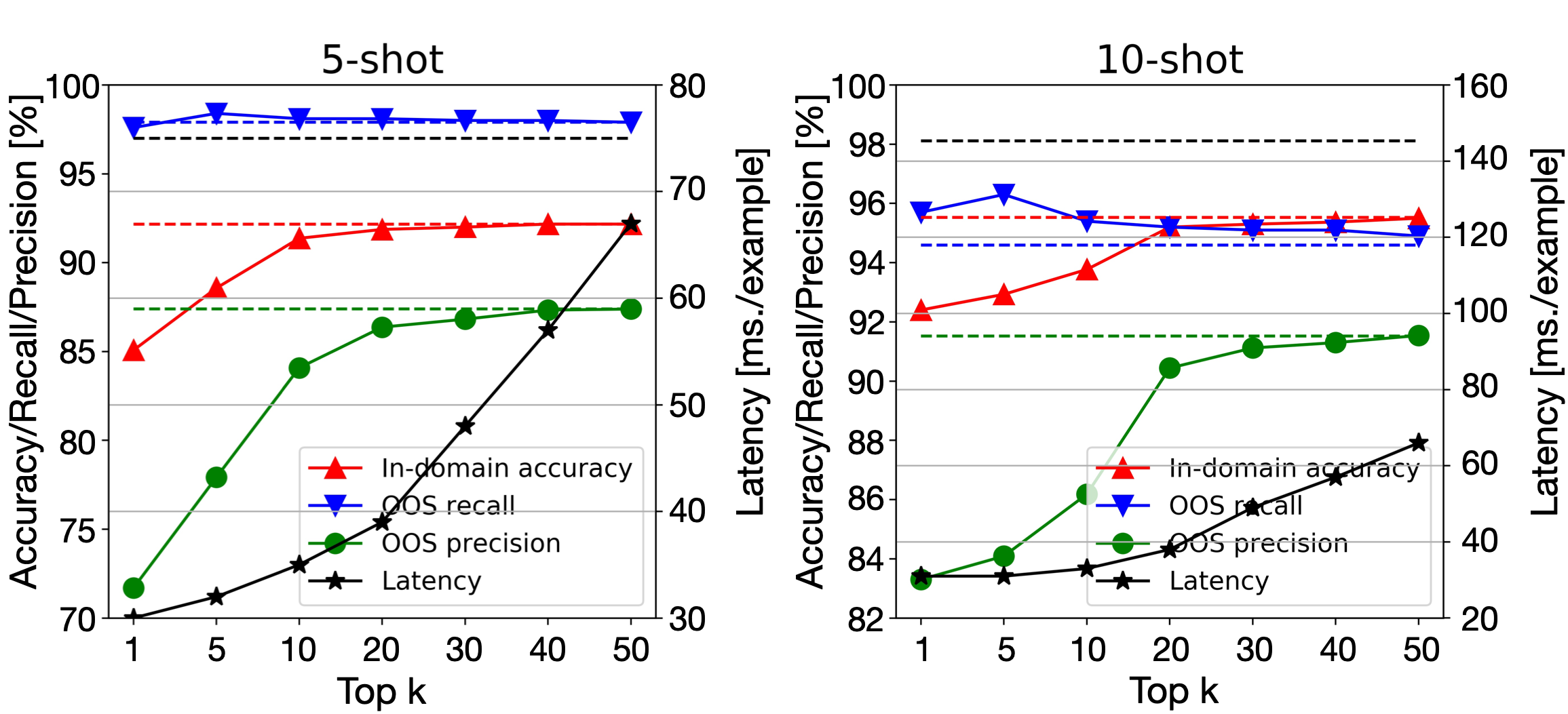}
    \end{center}
\caption{5-shot and 10-shot DNNC-joint development results on the banking domain, where the dash lines are DNNC results.}
\label{fig:joint_nli-appendix}
\end{figure*}

\paragraph{Faster inference}\label{appendix-DNNC-joint}
Figure~\ref{fig:joint_nli-appendix} shows the same curves in Figure~\ref{fig:join-nli} also for the 10-shot setting.
We can see the same trend with the 10-shot results.

\paragraph{Case studies}\label{Case Study}
Table~\ref{table:case-study} shows four DNNC prediction examples from the development set of the banking domain.
For the first example, the input utterance is correctly predicted with a high confidence score, and it has a similarly matched utterance to the input utterance;
for the second example, the input utterance is predicted incorrectly with a high confidence score, where the matched utterance is related to money but it has a slightly different meaning with the input utterance.
For the third example, the model gives a very low confidence score to predict an OOS user utterance as an in-domain intent; the last example is an incorrect case where the input utterance and the matched utterance have a topically similar meaning, resulting in a high confidence score for the wrong label, ``bill due.''
Based on these observations, it is an important direction to improve the model's robustness (even with the large-scale pre-trained models) towards such confusing cases.

\end{document}